\documentclass{article}

\usepackage[preprint]{corl_2025} 

\usepackage{graphicx} 
\usepackage{float}    
\usepackage{amsmath} 
\usepackage{amssymb}  
\usepackage{caption}
\usepackage[most, breakable]{tcolorbox}
\usepackage{lipsum} 
\usepackage[inline]{enumitem}
\usepackage{graphicx}
\usepackage{algorithm}
\usepackage{algpseudocode}
\usepackage{multirow}
\usepackage{makecell}
\usepackage{enumitem}
\usepackage{makecell}
\usepackage{caption}
\usepackage{xspace}
\usepackage{array} 
\usepackage{booktabs}
\usepackage{wrapfig}
\usepackage[most]{tcolorbox}

\usepackage{xcolor}
\usepackage{listings}
\usepackage[frozencache,cachedir=.]{minted}

\usepackage{tabularx}
\renewcommand{\arraystretch}{1.3}
\usepackage{parskip}  
\usepackage{makecell}
\usepackage{pifont} 
\newcommand{\tick}{\checkmark}
\newcommand{\cross}{\texttimes}

\definecolor{deepblue}{HTML}{0056FF}
\definecolor{softblue}{HTML}{A4C8FF}
\definecolor{coolcyan}{HTML}{7FFFD4}
\definecolor{sunnyyellow}{HTML}{FFEB3B}
\definecolor{warmorange}{HTML}{FF9800}
\definecolor{peachpink}{HTML}{FFCDD2}
\definecolor{softpurple}{HTML}{CE93D8}
\definecolor{forestgreen}{HTML}{388E3C}
\definecolor{midgray}{HTML}{9E9E9E}
\definecolor{charcoal}{HTML}{424242}
\definecolor{rosewood}{HTML}{65000B}
\definecolor{nightblue}{HTML}{1A237E}

\newtcolorbox[auto counter]{promptbox}[2][]{
  enhanced,
  breakable,
  enhanced jigsaw,
  colback=gray!5,
  colframe=black!40,
  sharp corners=south,
  coltitle=white,
  colbacktitle=black,
  title={Prompt~\thetcbcounter: #2},
  fonttitle=\bfseries\large,
  toptitle=5pt,
  bottomtitle=5pt,
  boxed title style={
    colframe=black,
    colback=black,
    sharp corners=south,
    size=small,
    boxrule=0pt
  },
  #1
}

\newcommand{\myalg}{{ApBot}\xspace}
\newcommand{\myalgbr}{{ApBot}\xspace}

\newcommand{\states}{S}
\newcommand{\gstates}{\states_g}
\newcommand{\acts}{A}
\newcommand{\macroaction}{\Phi}
\newcommand{\actseq}{{\bf a_\phi}}
\newcommand{\actgo}{\acts_{g}}
\newcommand{\actnb}{\acts_{n}}
\newcommand{\trans}{\mathcal{T}}
\newcommand{\transgo}{\trans_{g}}
\newcommand{\transnb}{\trans_{n}}

\newcommand{\astates}{\overline{\states}}
\newcommand{\agstates}{\astates_g}
\newcommand{\amacroaction}{\overline{\macroaction}}
\newcommand{\aacts}{\overline{\acts}}
\newcommand{\aactgo}{\aacts_{g}}
\newcommand{\aactnb}{\aacts_{n}}
\newcommand{\atrans}{\overline{\trans}}
\newcommand{\atransgo}{\atrans_{g}}
\newcommand{\atransnb}{\atrans_{n}}

\newcommand{\numbox}{N_b}

\newcommand{\allbox}{{\bf B}}

\newcommand{\secref}[1]{Sec. \ref{#1}}
\newcommand{\figref}[1]{Fig. \ref{#1}}
\newcommand{\tabref}[1]{Table \ref{#1}}
\newcommand{\apref}[1]{Appendix \ref{#1}}

\newcommand{\reduceteaserinterval}{\vspace{0pt}}
\newcommand{\reducesecinterval}{\vspace{0pt}}
\newcommand{\reduceparagraphinterval}{\vspace{0pt}}

\newcommand{\cotimage}{LLM as policy w/ image\xspace}
\newcommand{\cotaction}{LLM as policy w/ grounded actions\xspace}
\newcommand{\mynoreason}{\myalg w/o button policy\xspace}
\newcommand{\mynomodel}{\myalg w/o model\xspace}
\newcommand{\mynoupdate}{\myalg w/o close-loop update\xspace}

\title{
Robot Operation of Home Appliances\\by Reading User Manuals
} 

\author{
  Jian Zhang,  Hanbo Zhang,  Anxing Xiao, 
 David Hsu\\
  School of Computing \& Smart System Institute\\
  National University of Singapore\\
corresponding to \texttt{zhang.jian@u.nus.edu} \\
}

\begin{document}
\maketitle


 \begin{figure}[h]
     \centering
     \includegraphics[width=1.0\linewidth]{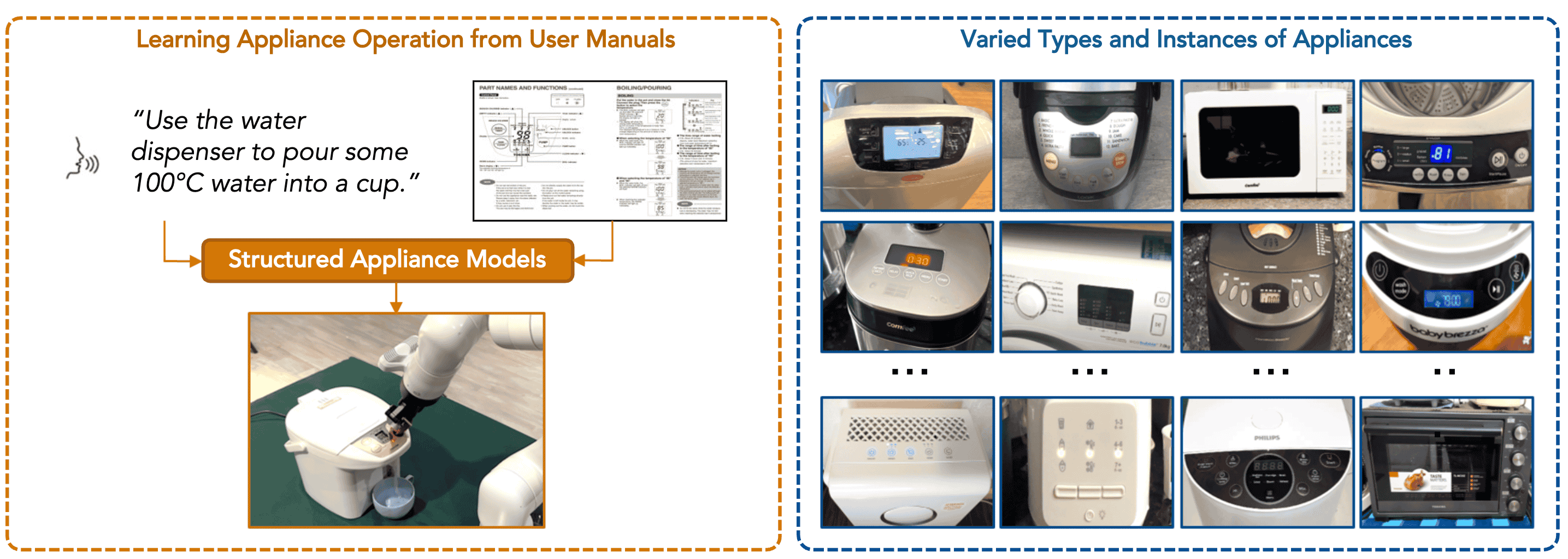}
     \caption{
\myalg enables robots to operate \textbf{diverse}, \textbf{novel} household appliances from natural language instructions with manuals and visual observations in a zero-shot manner. It follows open-ended instructions to generate grounded multi-step actions for complex appliance operation.
}
     \label{fig:main_figure}
 \end{figure}

\reduceteaserinterval
\begin{abstract}
    Operating home appliances, among the most common tools in every household, is a critical capability for assistive home robots. This paper presents \myalgbr, a robot system that operates \textit{novel} household appliances by ``reading'' their user manuals. \myalgbr faces multiple challenges: (i) infer goal-conditioned partial policies from their unstructured, textual descriptions in a user manual document, (ii) ground the policies to the appliance in the physical world, and (iii) execute the policies reliably over potentially many steps, despite compounding errors.  To tackle these challenges, \myalgbr constructs a structured, symbolic model of an appliance from its manual, with the help of a large vision-language model (VLM). 
    It grounds the symbolic actions visually to control panel elements. Finally, \myalgbr closes the loop by updating the model based on visual feedback. 
    Our experiments show that across a wide range of simulated and real-world appliances,  ApBot achieves consistent and statistically significant improvements in task success rate, compared with state-of-the-art large VLMs used directly as control policies. These results suggest that a structured internal representations plays an important role in robust robot operation of home appliances, especially,  complex ones.
\end{abstract}


\section{Introduction}
\reducesecinterval
Operating household appliances is a fundamental yet underexplored topic for assistive robots at home. It can dramatically extend the capabilities of robots in daily environments. Unlike passive tools, appliances encapsulate complex, high-level functionalities to change object states (e.g., cooking, cleaning, heating) that no direct manipulation skills can replicate. Ideally, robots are expected to perform general-purpose appliance operations automatically with the help of manuals. However, this task poses several crucial challenges to existing robotic systems. First, manuals are usually unstructured documents of text and symbols, making it difficult for robots to understand. Furthermore, appliances are designed to follow constrained workflows governed by operation modes, making them less error-tolerant. Therefore, we aim to answer the following question: \textit{How can we enable robots to generate visually grounded policies for novel appliance operation with user manuals?}

\reduceparagraphinterval

To this end, we propose \myalg, a generalizable open-world framework for appliance operation (see Figure~\ref{fig:main_figure}). With the help of large vision-language models (LVLMs), \myalgbr constructs a symbolically structured model of novel appliances conditioned on user manuals for interpretable, controllable, and verifiable policy generation. 
\myalg constructs structured appliance models of novel appliances, i.e., state machines, using the user manual and visual observation. The model captures symbolic representations of executable actions, appliance states, and transition rules. 
Given a natural language task, the model generates an action sequence grounded in the panel layout, which is then executed by a low-level skill primitive.

To account for the inherent ambiguity of the manuals and open-world issues \cite{jiang2019open} of novel appliances, \myalgbr iteratively updates the models based on execution outcomes to align the model with real world appliance behavior. As a result, \myalgbr can effectively operate novel, complex appliances robustly with language instructions without further training.

\reduceparagraphinterval

Structured representation, along with online error correction, significantly improves robustness, especially against complex home appliances. To evaluate \myalgbr, we constructed from the manuals \cite{manualslib, internetarchive} a simulated benchmark containing 6 home appliance types and 30 interactive instances, including \textit{dehumidifier}, \textit{bottle washer}, \textit{rice cooker}, \textit{microwave oven}, \textit{bread maker}, and \textit{washing machine}.
For each appliance, we design $10$ different natural language instructions. Since appliances vary in complexity, we standardize the number of variables (e.g., \textit{time} and \textit{temperature}) used in instructions for each type, ranging from easy to hard. We compare \myalgbr with methods purely based upon LLMs or VLMs \cite{wei2022chain, deitke2024molmo, hurst2024gpt}. Results demonstrate that our method significantly and consistently outperforms them with a clear margin, illustrating its effectiveness and robustness for complex home appliance operation. 
We also deploy our system in the real world with a Kinova Gen3 arm and demonstrate the effectiveness of our proposed method on multiple real-world appliances.

\reduceparagraphinterval

In summary, our main contribution is a novel, symbolically structured representation for generalizable household appliance operation. It bridges gaps between unstructured inputs and policies, hence enabling robots to make controllable decisions and achieve reliable performance for the operation of novel appliances with visual input by reading user manuals. 

\reducesecinterval
\section{Related Work}
\reducesecinterval
\paragraph{Robotic Appliance Operation}
Previous work related to robotic appliance operation falls into three categories: perception, low-level manipulation, and long-horizon plan generation.
For perception, existing works mainly concentrate on button localization, by leveraging edge-based visual features \cite{wang2010robot, abdulla2016robust}, RFID tags \cite{nguyen2009pps}, or neural networks \cite{zhu2018novel, liu2021large, yuguchi2022toward, verzic2024recovering}.

For low-level skills, button manipulation is important, hence, many works focus on physical interaction with diverse button types by specializing fingertips \cite{wang2018robot} or leveraging additional sensory feedback \cite{sukhoy2010learning, wang2018robot}.

Long-horizon appliance operation remains relatively underexplored. A recent work \cite{yuguchi2022toward} uses handcrafted behaviors to generate plans for home appliance operation, but lacks generalization to novel appliances or tasks.
By contrast, our approach learns to model the appliances by reading the user manuals, hence enabling operation policy generation for new appliances and task instructions in a zero-shot manner.

\reduceparagraphinterval
    
\paragraph{Foundation Models for Robotic Decision Making}
Foundation models have been widely applied to robotic decision-making \cite{yang2023foundation}. Prompting large models to generate structured representations, such as logic frameworks \cite{liu2022lang2ltl, liu2023llm+, chen2024autotamp, vu2024coast} and codes \cite{lin2023text2motion, liang2023code, singh2023progprompt, schick2023toolformer, yin2024context, yao2022react} enables integration with external solvers or executors, though often under strict syntax constraints. 
It has been shown to improve reasoning precision at the cost of generality, particularly for long-horizon tasks with hard constraints \cite{chen2023llm, shinn2023reflexion, song2023llm, nottingham2023selective}. Unstructured representations such as textual reasoning trees \cite{jiang2024roboexp, ao2024llm, zhou2024llm, chen2024integrating} or natural languages \cite{rana2023sayplan, ahn2022can, huang2022inner, raman2024cape, wang2024llm} offer flexibility and generality but suffer from ambiguity and lack correctness guarantees. Approaches such as syntax validation \cite{lin2023towards, lin2022planning}, corrective feedback \cite{raman2022planning, raman2024cape}, and prompt optimization \cite{lu2024fine} aim to take advantage of both by modular design. 

Particularly, domain-specific languages (DSLs) can enhance LVLM reliability and robustness via in-context learning \cite{zhang2024dkprompt, zheng2024fine, singh2023progprompt, wang2024grammar}, offering key insights for our structured design for appliances. 
We adopt a DSL-based design tailored to appliance operation and generate it via in-context learning and syntax validation. To support numerical computation, we enable the invocation of external function calls similar to \cite{liang2023code}. This ensures generalizability while maintaining correctness guarantees.

\reduceparagraphinterval

\paragraph{Graphical User Interface Agents} 
Graphical User Interface (GUI) agents \cite{nguyen2024gui, wang2024gui}, similar but unlike appliance operation, operate devices through software interfaces. 

Recently, the prevailing approach involves fine-tuning VLMs on datasets of web or mobile interactions, such as clicking, typing, or tapping, to directly predict on-screen actions \cite{lispotlight, cheng2024seeclick, wu2024atlas, gou2024navigating}. 

An alternative line of work uses visual prompting, where VLMs select actions based on overlaid marks (e.g., boxes or numbers) on GUI screenshots \cite{lu2024omniparser, he2024webvoyager, koh2024visualwebarena, qin2025ui}. 
Yet, these methods often struggle with complex GUIs, especially when the documentation is incomplete or unstructured.
Another line of related research is Retrieval-Augmented Generation (RAG) \cite{lewis2020retrieval}, which augments model input with information retrieval from external sources, such as online manuals or documents. Building on this idea, we introduce structurally grounded representations for appliance operation, which encode knowledge from user manuals to enable robust plan generation.

\reducesecinterval
\section{\myalg}
\reduceparagraphinterval

 \begin{figure}[t]
     \centering
     \includegraphics[width=1.0\linewidth]{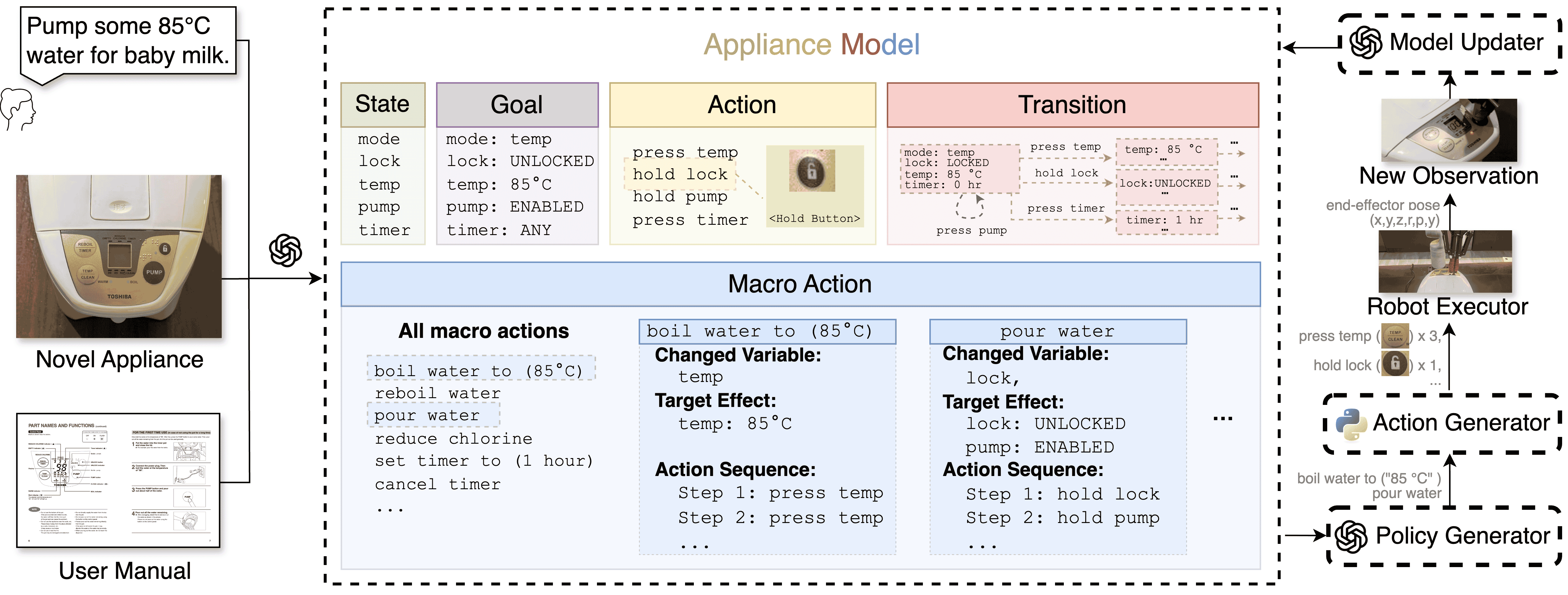}
     \vspace{-10pt}
     \caption{Overview of \myalg. The key of \myalg is the appliance model built from a textual manual, based on which we generate actions for novel appliance operation. To handle possible model errors, we calibrate the model with observable feedback during execution in a close-loop manner.}
     \label{fig:model_pipeline}
     \vspace{-15pt}
 \end{figure}

\subsection{Problem Formulation}
\label{sec:problem-formulation}
\reduceparagraphinterval
We consider the task of household appliance operation based on parameterized natural language instructions and visual observations, with the help of textual information from user manuals. Given a natural language instruction (e.g., \textit{turn on the device and set the Cook Rice mode}), the system must interpret the goal state, reason about appliance constraints, and generate a sequence of executable low-level actions that complete the task.

\reduceparagraphinterval
We formulate the appliance as a state machine \cite{brand1983communicating} with tasks specified as a subset of goal states: $\mathcal{M}=\langle \states, \acts, \trans, \gstates \rangle$. Each state $s$ in $\states$ comprises a list of variables for the appliance (e.g., \texttt{on} or \texttt{off}, \texttt{time}, \texttt{laundry program}, \texttt{temperature}). 

For action space $\acts$, we empirically find that most actions of operating a home appliance can be categorized into two classes: $\actgo$ and $\actnb$. Any action in $\actgo$ will directly \textit{go} to a pre-defined, specific state (e.g., ``Menu" to enter menu mode), if the current state is in a specific subset of $S$ (e.g. power on, child lock disabled). Any action in $\actnb$ (e.g., ``+" and ``-") will turn the state of a variable to a neighborhood value, subject to the current state. Accordingly, we have a deterministic transition model for each type $s'=\trans(s,a)$, where $\trans\in \{\transgo, \transnb\}$. Given the above formulation, we aim to find a shortest sequence of actions $\textbf{a}^*=(a_1^*, a_2^*, ..., a_T^*)$ to achieve an arbitrary goal state in $\gstates$, a subset of $\states$.

\reduceparagraphinterval
However, in practice, we cannot have access to the underlying true model $\mathcal{M}$. Instead, we construct an approximated one $\overline{\mathcal{M}}=\langle \astates, \aacts, \atrans, \agstates \rangle$ from the manual, upon which we generate the operation policy for the appliance. To construct $\overline{\mathcal{M}}$, we assume that robots can observe the home appliances visually and have the corresponding manuals in raw text. We assume that the users will interact with robots in natural languages to specify tasks, which defines $\agstates$ given $\astates$. We posit that LVLMs can read raw textual information in the manual and build a partially correct appliance model $\overline{\mathcal{M}}$ accordingly, by proper design.

\reduceparagraphinterval
We propose a system, \myalg, for natural language control of household appliances by combining user manuals and visual input. As shown in Figure~\ref{fig:model_pipeline}, the system (1) constructs a symbolic model from manuals (Sec.~\ref{sec:model_construct}), (2) grounds symbolic actions to visual control elements (Sec.~\ref{sec:action-grounding}), (3) models transitions as macro actions (Sec.~\ref{sec:macro-action}), and (4) executes tasks with closed-loop updates based on real-time feedback (Sec.~\ref{sec:online-update}) to address errors of model construction.

\reducesecinterval
\subsection{Construct Structured Appliance Model}
\label{sec:model_construct}
\reduceparagraphinterval

\paragraph{Modelling States, Actions, Transitions.}
To construct the symbolic model $\overline{\mathcal{M}}=\langle \astates, \aacts, \atrans, \agstates \rangle$, we sequentially generate the state space $\astates$, action space $\aacts$, and transition function $\atrans$ with the help of LVLM agents \citep{hurst2024gpt} using prompting \cite{wei2022chain}. Concretely, for each of them, we provide the manual, a predefined output format, and a complete list of valid options, and ask LVLMs to generate the appliance model accordingly. Model generation goes in an in-context way \cite{brown2020language}, where examples are provided in the input to improve robustness. 
 
Syntax checkers are then applied to ensure output validity, with up to three regeneration attempts if errors occur, e.g, violations of constraints.
We provide a detailed example of appliance modeling in \apref{ap:appliance_model}, along with the corresponding prompts in \apref{ap:prompts}, and list the syntax checks used for validation in \apref{ap:syntax_checker}.

\reduceparagraphinterval
\paragraph{Extracting Goals from Instructions.}
To infer the goal state $\agstates$ from a natural language instruction, we prompt the LVLM agent to produce a partial assignment over symbolic variables that fulfills the task requirements. For example, given the instruction “set cooking power to high and cooking time to 2 minutes”, the inferred goal corresponds to a symbolic state where \texttt{power} is \texttt{high} and \texttt{time} is \texttt{2 minutes}, while all other variables remain unconstrained.

\reducesecinterval
\subsection{Action Grounding}
\label{sec:action-grounding}

\begin{wrapfigure}{R}{0.45\textwidth} 
  \centering
  \vspace{-10pt} 
  \includegraphics[width=0.43\textwidth]{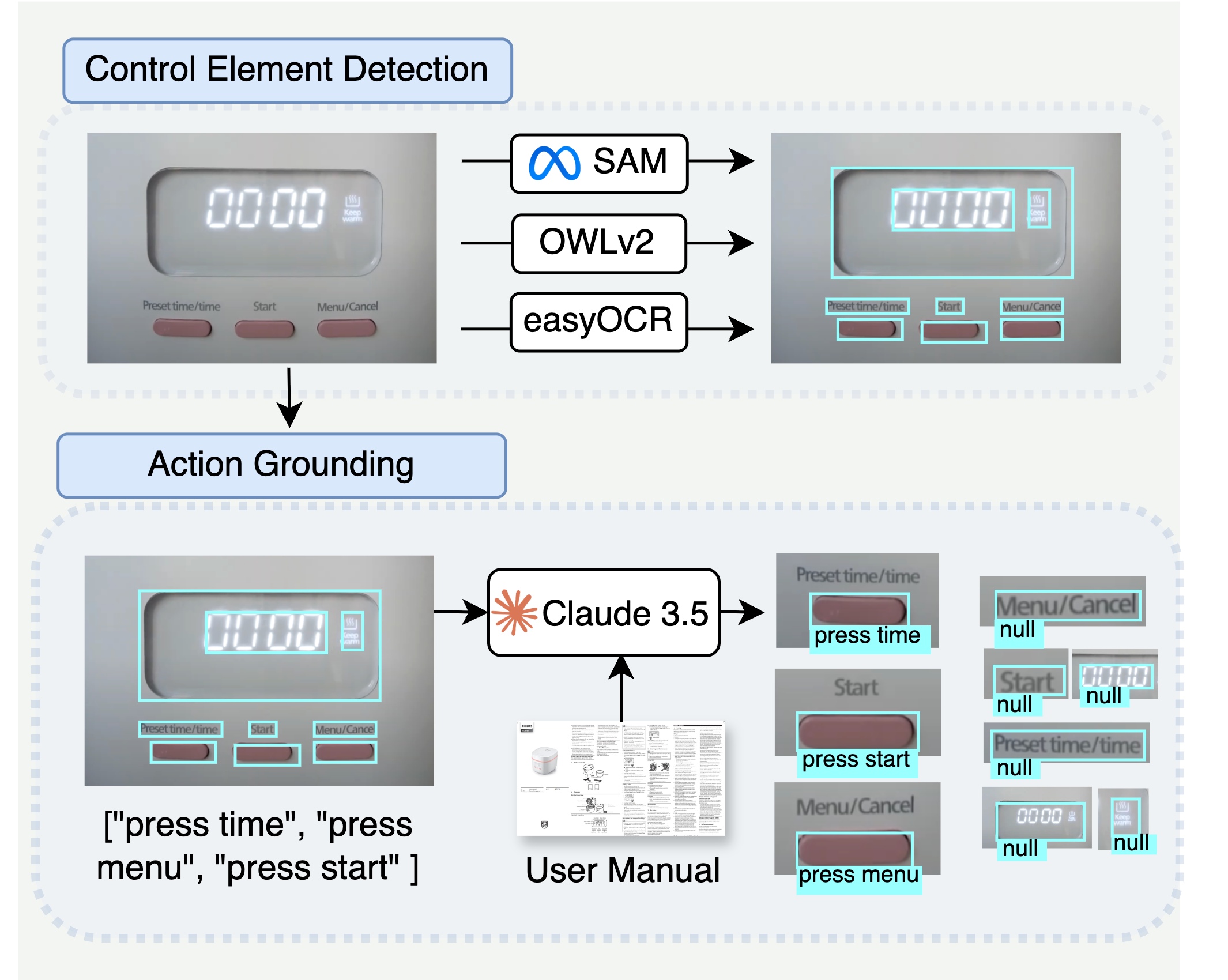}
  \caption{Overview of action grounding with visual observations.}
  \vspace{-0.7cm}
  \label{fig:overview_grounding}
\end{wrapfigure}

\reduceparagraphinterval
To make actions physically executable, we need to ground the symbolic actions visually onto the observed \textit{control panel elements}, which are interactive components of an appliance, such as buttons, dials, and printed touch pads. Each grounded action $\hat{a}$ is a tuple $\hat{a} = (a, b, \sigma)$, where $a\in\aacts$ is the symbolic action from the manual, $b$ is a bounding box of the visual region, and $\sigma \in \{\texttt{press}, \texttt{hold}, \texttt{turn}\}$ denotes the primitive robot skill required to execute the action.
We demonstrate the pipeline of action grounding in \figref{fig:overview_grounding}.

\reduceparagraphinterval
\paragraph{Control Element Detection.} We assume the control panel elements can be clearly detected using existing object detectors. We prioritize high recall in detection to ensure all buttons are captured. To this end, we run three models in parallel. Segment Anything (SAM) \cite{kirillov2023segment} segments the image into regions of visually distinct entities. OWL-ViT2 \cite{minderer2023scaling} is queried with prompts of “button”, “dial”, and “switch” to detect control elements. An OCR model \cite{easyocr} extracts regions of visible text labels. We take the union to form a candidate set for all control elements. To remove false positives, we sort the bounding boxes in descending order of detection confidence. For each pair of boxes $(b_i, b_j)$, if $\text{IoU}(b_i, b_j) > 0.85$, we discard the box with lower detection confidence, as overlapping boxes likely refer to the same object. We further use LVLMs to check whether each remaining box likely contains a valid control panel element, following \cite{yang2023set}. So far, we have a set of boxes $\allbox = \{b_i\}_{i=1}^{\numbox}$, where $\numbox$ is the number of boxes.

\reduceparagraphinterval
\paragraph{Actions Grounding.}
To get the executable action $\hat{a}=(a,b,\sigma)$, we need to do visual grounding, i.e., an injection from symbolic actions $\aacts$ to boxes $\allbox$, and identify the manipulation type $\sigma$ for each $a\in\aacts$. To do so, we first query LVLMs to assign an action $a\in\aacts$ for each $b\in\allbox$, i.e., a mapping from $\allbox$ to $\aacts$. Inversely, now, each $a$ may: (1) have a unique box $b$; (2) have no box; (3) have a set of boxes $\allbox_a\subseteq\allbox$, $|\allbox_a|>1$. For (1), it is ideal. For (2), we directly remove it from $\aacts$ since it is no longer executable, hence, all tasks involving this action will fail. For (3), we impose the following two heuristics by further prompting the LVLMs: 1. the box including clear physical boundaries is preferred; 2. the box with an icon-based label is preferred over text-only ones. Finally, we assign the manipulation type $\sigma$ directly based on the description of $a$ along with the assigned visual region $b$, forming $\hat{a}=(a,b,\sigma)$, which can be executed using the corresponding low-level primitive skill.

\reducesecinterval
\subsection{Structured Transition Modeling with Macro Actions}
\label{sec:macro-action}

\reduceparagraphinterval
Perfectly modeling appliances with one shot is impractical due to the inherent ambiguity of the manuals (e.g., ``adjust to desired level") and inevitable errors of LVLMs (e.g., hallucinations), making planning hard. Instead, we leverage two observations of common appliance designs: 

(1) variable adjustments often follow consistent action sequences, such as pressing “+” repeatedly or entering digits conditioned on an explicitly specified number string; and (2) user manuals often describe step-by-step tutorials for common usages. We formalize user manuals as macro actions $\amacroaction$, a constructed version of the underlying ground truth set of macro actions $\macroaction$, for efficient and generalizable appliance modeling.

\reduceparagraphinterval
\paragraph{Definition of Macro Actions.} Macro action $\phi\in\macroaction$ is a parameterized sequence of symbolic actions that encapsulate a meaningful functionality (e.g., \texttt{PowerOn}, \texttt{SetCookingMenu}). Specifically, each macro action $\phi$ consists of a symbolic action sequence $\actseq=(a_1,a_2,...)$, where $a_i\in\acts$. Besides, each macro action has a set of variables ${s_v}$ on which it imposes effects and a macro transition describing these effects $\Gamma(s_v)$. Note that $s_v$ is part of the full state $s$. Formally, $\phi=(s_v,\Gamma(s_v),\actseq)$. For example, the macro action $\phi=\texttt{SetCookingMenu(LongGrain)}$ may include two symbolic actions: $a_1=\texttt{press\_menu}$ and $a_2=\texttt{turn\_dial}$, 
which will change the values of variables $s_v=(\texttt{appliance mode},\texttt{cooking menu})$, so as to (1) enter the cooking menu setting mode; and (2) set the cooking menu to \texttt{LongGrain}. Macro actions enable LVLMs to plan by specifying high-level subgoals, simplifying reasoning by substantially reducing the reasoning horizon. 

\reduceparagraphinterval
\paragraph{Modeling Macro Actions.}
The full list of macro actions, including the corresponding variables and target effects, is extracted by prompting LVLMs (see \apref{ap:prompts}).
To fully model the macro actions, we need to generate a sequence of low-level actions $\actseq$ based on the current variable $s_v$ and target transition $\Gamma(s_v)$.
Specifically, $\actseq$ can be directly computed from two transition types: $\atransnb$ for $\actnb$ actions like "+" or "-", and $\atransgo$ for go-to actions in $\aactgo$. The computation invokes codes directly, similar to \cite{liang2023code}, based on the current and goal values. 

For actions in $\actnb$, we compute the number of repeats based on its transition model (e.g., from 1 to 5 requires 4 presses of ``+").
For actions in $\actgo$, we translate the description in the manual directly to get the specific actions.

\reducesecinterval
\subsection{Closed-loop Task Execution}
\label{sec:online-update}

\reduceparagraphinterval
\paragraph{Automatic Execution from Macro Actions.}
To specify the executable actions for robots, \myalg generates a parameterized symbolic task policy $\pi = [\phi_1, \phi_2, ..., \phi_T]$ conditioned on the inferred goal $\agstates$, where each $\phi_i \in \amacroaction$ is a parameterized macro action, covering one or more variables specified in the goal state. To do so, we feed the list of applicable macro actions to LVLMs, along with the textual specification of $\agstates$, and generate the policy $\pi$ directly. Actions in each $\phi_i$ are determined via its transition rule $\atrans_i$.
The robot executes low-level actions sequentially through a set of parameterized primitive skills. Implementation details of the primitive skills used on the real robot are provided in Appendix \ref{ap:realworld}.

\reduceparagraphinterval
\paragraph{State Estimation and Model Updates.}
To ensure robustness against inaccuracies in generated appliance models, we adopt closed-loop model calibration. After each macro action is executed, the system gets new observations and tracks the state to check whether the target state is achieved. In simulation, the environment returns a textual description (e.g., ``\textit{time is set to 30 min}"), while in the real world, an image is captured as visual feedback. In both cases, the feedback is passed to LVLMs to infer the actual value of the corresponding variables. If the actual value fails to match the expected one by the transition, \myalg first traverses the full value range of the variable to observe how it responds to repeated actions. \myalg then utilizes the executed trace to update the transition $\atrans$ (e.g., step size, value bounds), based on which the action sequence in this macro action $\actseq$ will be updated accordingly. With the updated macro actions, a new plan will be regenerated. For example, if \texttt{press("+")} produces a sequence like $0,\text{min} \rightarrow 10,\text{min} \rightarrow \dots \rightarrow 60,\text{min} \rightarrow 0,\text{min}$, the updated rule reflects a 10-minute step and a wraparound at 60 minutes. The details of state tracking and model updates are elaborated in \apref{ap:model_updates}.

\reducesecinterval
\section{Experiments}
\reducesecinterval

We evaluate \myalgbr by answering three questions: \textit{(1) How does it compare to state-of-the-art LVLM agents for home appliance operation? (2) What are the main contributors of \myalg? (3) How does it perform on real-world appliances?} For (1), we compared \myalgbr with leading LVLM agents using unstructured inputs and found that \myalgbr consistently achieves higher success rates with fewer steps in both simulation and real-world settings. For (2), ablations show that structured appliance models, structured reasoning, and closed-loop updates are all critical for robust operation. 
For (3), real-world deployments demonstrate the effectiveness of \myalgbr on unseen appliances and long-horizon tasks.

\reducesecinterval
\subsection{Experimental Settings}

\reduceparagraphinterval
\paragraph{Evaluation Benchmark.} Our evaluation aims to systematically assess the effectiveness and generalization of \myalgbr across varying appliances. We construct a simulated benchmark of 30 interactive appliances with their manuals across 6 categories.

\begin{figure*}[tb]
  \centering
  \begin{tabular}{@{}cccccc@{}}
    \hspace{0.1cm}
    \footnotesize{Induction Cooker}
    &
    \multicolumn{5}{l}{ Instruction: \textit{Select the HotPot mode and adjust the power setting to 2000 W.}}
    \\
    \includegraphics[width = 0.179\textwidth]{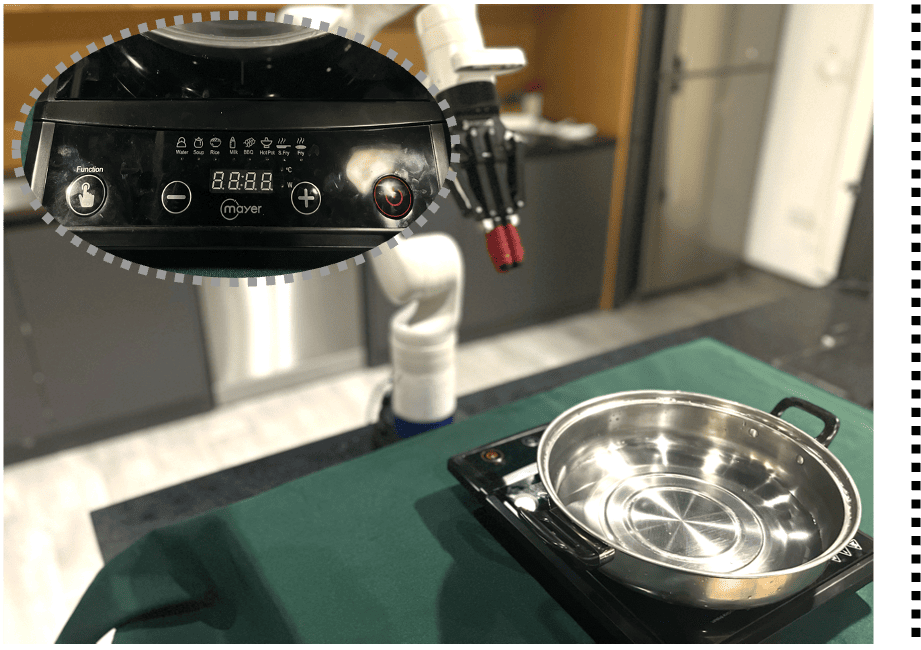} \hspace{-0.35cm}
    &
    \includegraphics[width = 0.148\textwidth]{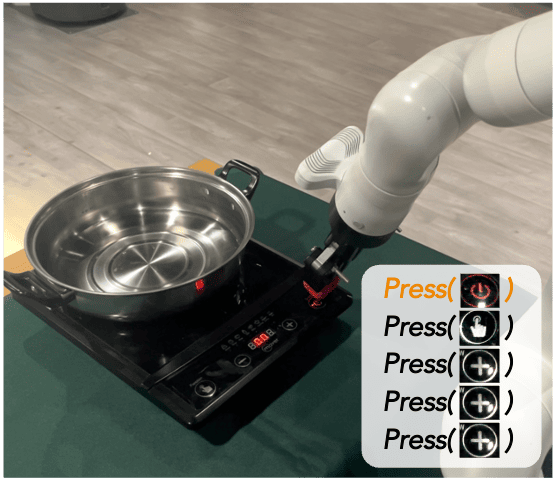} \hspace{-0.35cm}
    &
    \includegraphics[width = 0.148\textwidth]{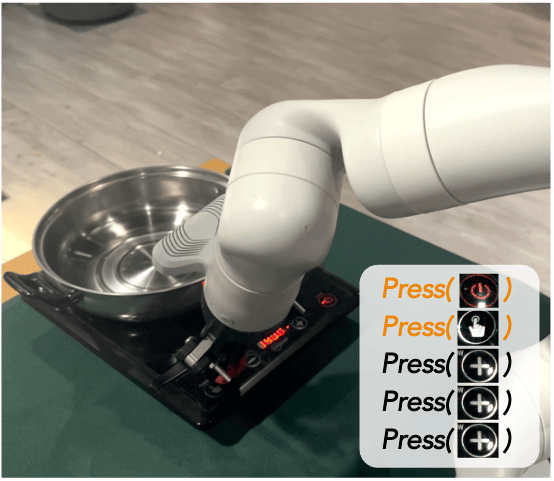} \hspace{-0.35cm}
    &
    \includegraphics[width = 0.148\textwidth]{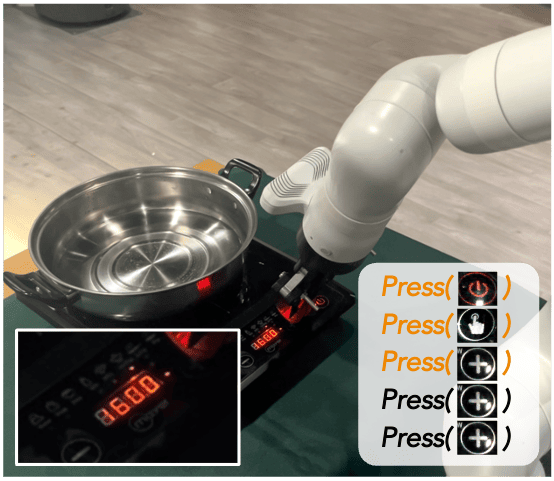} \hspace{-0.35cm}
    &    
    \includegraphics[width = 0.148\textwidth]{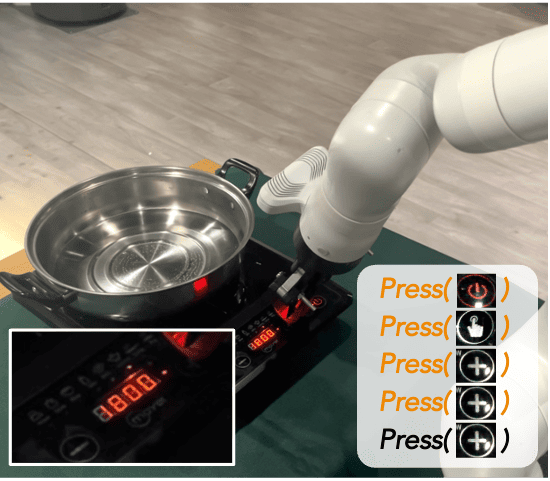} 
    \hspace{-0.35cm}
    &    
    \includegraphics[width = 0.148\textwidth]{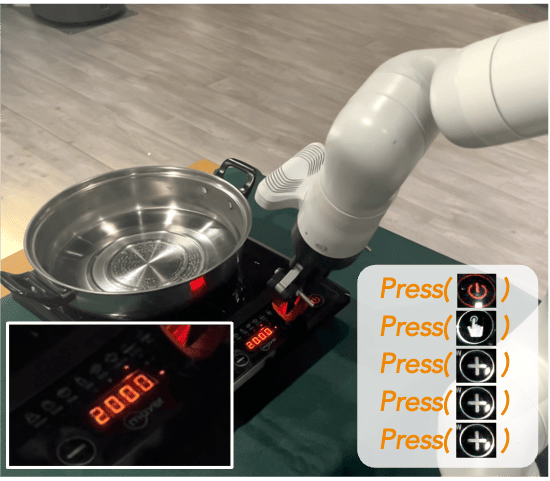} 
    \\    
    \hspace{0.1cm}
    \footnotesize{Water Dispenser}
    &
    \multicolumn{5}{l}{ Instruction: \textit{Set temperature to 60 degrees and then pour water.}}    
    \\
    \includegraphics[width = 0.179\textwidth]{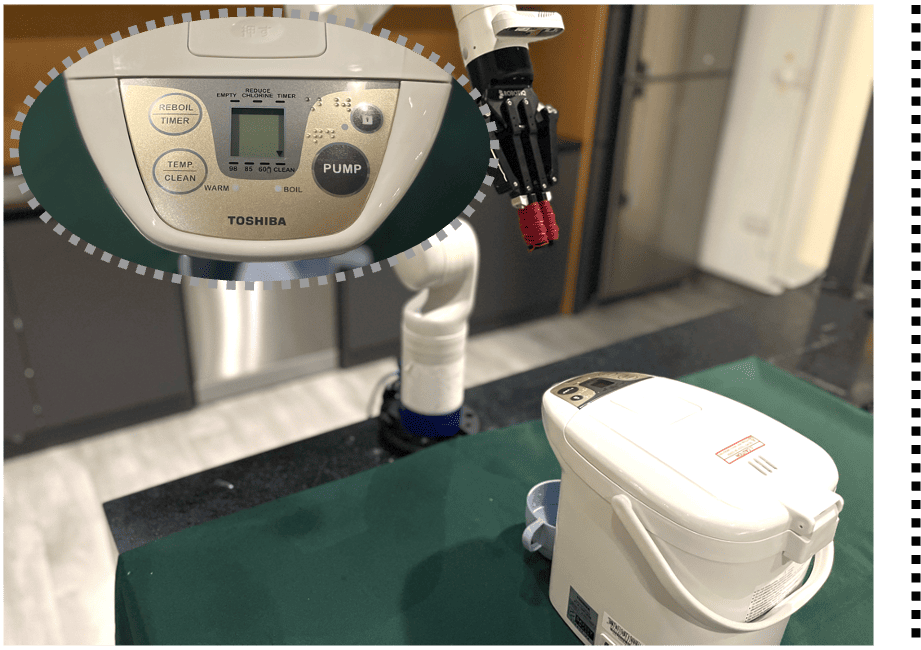} \hspace{-0.35cm}
    &
    \includegraphics[width = 0.148\textwidth]{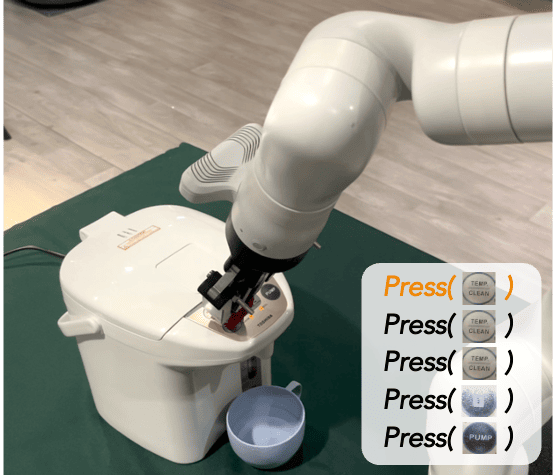} \hspace{-0.35cm}
    &
    \includegraphics[width = 0.148\textwidth]{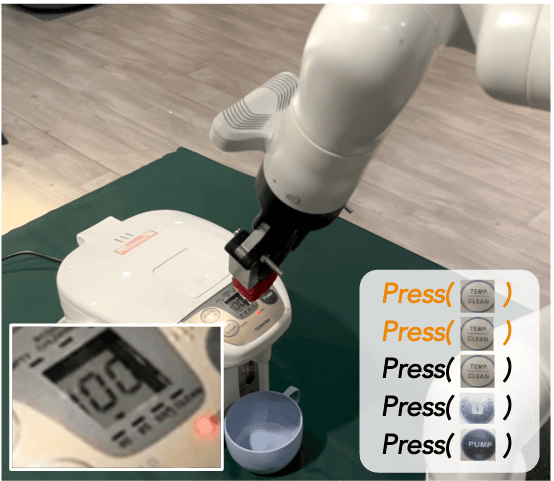} \hspace{-0.35cm}
    &
    \includegraphics[width = 0.148\textwidth]{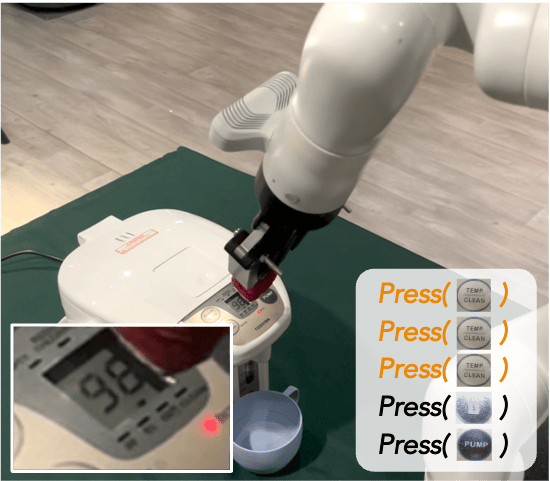} \hspace{-0.35cm}
    &    \includegraphics[width = 0.148\textwidth]{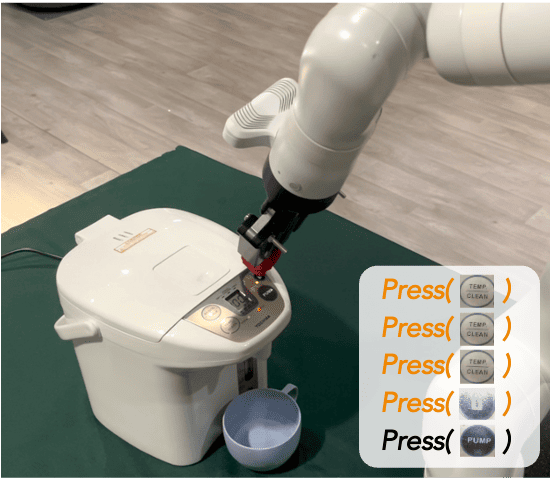} 
    \hspace{-0.35cm}
        &    \includegraphics[width = 0.148\textwidth]{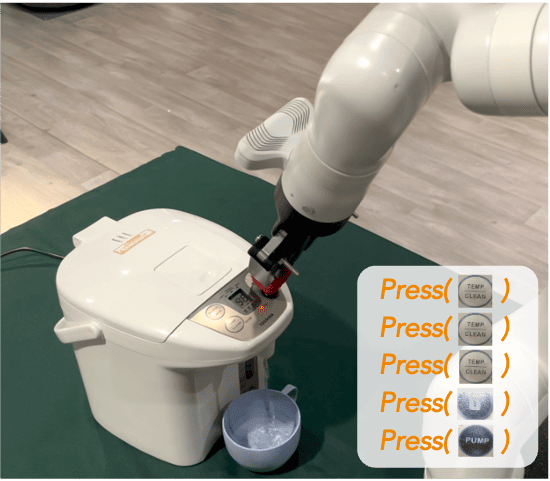} 
    
    \vspace{-0.2cm}

  \end{tabular}
  \caption{Snapshots of our system operating an induction cooker and a water dispenser.
}
\vspace{-0.6cm}
\label{fig:real_appliance_results}
\end{figure*}

Task instructions are designed with varying numbers of variables, from simple to hard.
Each instruction specifies explicit target values for the adjustable variables.
In total, we evaluated each method on a set of 300 goal-directed natural language instructions, 10 per appliance instance.
For automatic evaluation, each appliance in the benchmark is paired with a symbolic simulator that models true action effects and provides corresponding feedback to the algorithms. Full dataset including appliances image, user manual and instructions, along with the simulator details, is provided in Appendix~\ref{ap:eval_dataset}. For real-world evaluation, the system is deployed on three appliances using a Kinova Gen3 robot, following the same structured pipeline but relying on realistic visual observations for feedback. \figref{fig:real_appliance_results} shows the experimental setup.

\reduceparagraphinterval
\paragraph{Baselines.} We compare \myalg with several baselines designed to ablate key components. \textit{\cotimage} uses LVLMs for all modules, including visual grounding \cite{yang2023set} and reasoning based on unstructured, textual inputs. \textit{\cotaction} reasons over grounded actions from \secref{sec:action-grounding}. 
We also conduct ablations as follows.
\textit{\mynomodel} does not build appliance model $\overline{\mathcal{M}}$. Instead, LVLMs (1) decide which action to execute directly; (2) if the LVLM deems that repeating steps is required, it invokes codes to get the required action sequences. \textit{\mynoreason} builds a structured model $\overline{\mathcal{M}}$, and follows the macro actions in policy $\pi$ strictly, but relies on LVLMs for low-level action generation instead of leveraging the transition $\atrans$. \textit{\mynoupdate} disables model updates from observation feedback and executes in open-loop. Besides, we compare our action grounding approach with \textit{Molmo}, a state-of-the-art visual grounding method. We elaborate all model settings and baselines in Appendix~\ref{ap:experiment-setup}.

\reduceparagraphinterval
\paragraph{Evaluation metrics.}  Success is defined as achieving all specified values correctly. For metrics, we evaluate (1) \textit{Success Rate} within 25 reasoning steps, (2) \textit{Average Steps} taken before success or termination, and (3) \textit{Success weighted by Path Length (SPL)} to evaluate the weighted success rate considering the actual execution steps. Optimal steps are computed using oracle appliance models and task policies that specify the ground-truth action sequences.
\reducesecinterval
\subsection{Simulation Results}

\begin{figure}[t]
    \centering
    \includegraphics[width=\textwidth]{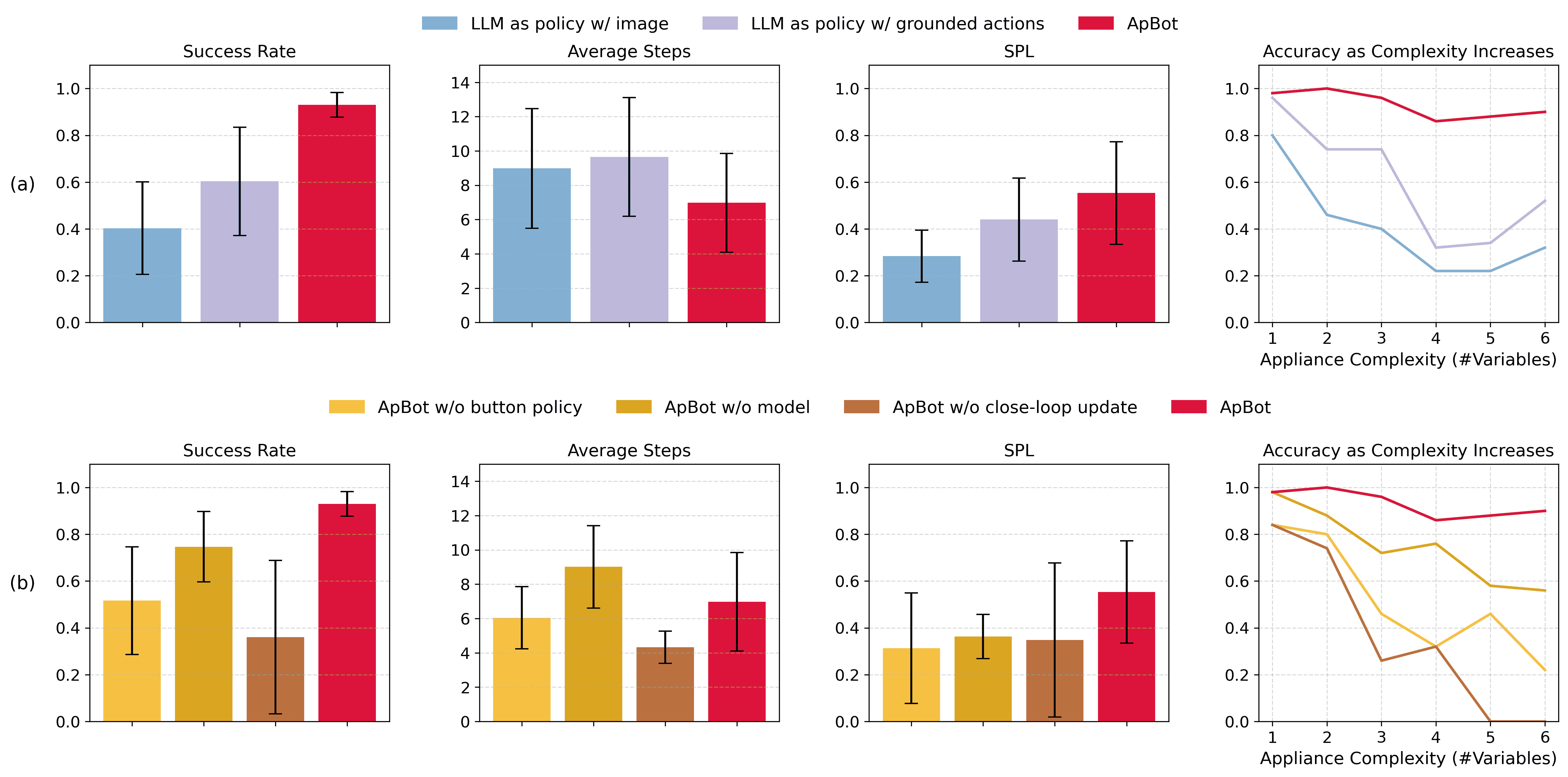}
    \vspace{-0.5cm}
    \caption{
        Overall performance of home appliance operation, including average task success rate (SR), average number of execution steps (Average Steps), and SPL (Success weighted by Path Length) across (a) baseline methods and (b) our ablations. Both performance and derivations are across appliance types.
    }
    \label{fig:overall_perform}
    \vspace{-0.5cm}
\end{figure}

\reduceparagraphinterval
\paragraph{How does our framework compare to large-scale vision-language agents?}
The overall performance of \myalgbr is shown in \figref{fig:overall_perform}. Compared to purely LVLM-based agents (\textit{\cotimage} and \textit{\cotaction}), \myalgbr achieves significantly better performance overall.
Noticeably, comparing \textit{\cotimage} and \textit{\cotaction}, visually grounded actions overall help for appliance operation tasks. This shows that current state-of-the-art LVLMs are not yet good at open-vocabulary detection or visual grounding tasks, especially those requiring fine-grained text recognition.
\begin{wrapfigure}{R}{0.37\textwidth}
  \centering
  \vspace{-1cm} 
  \includegraphics[width=0.36\textwidth]{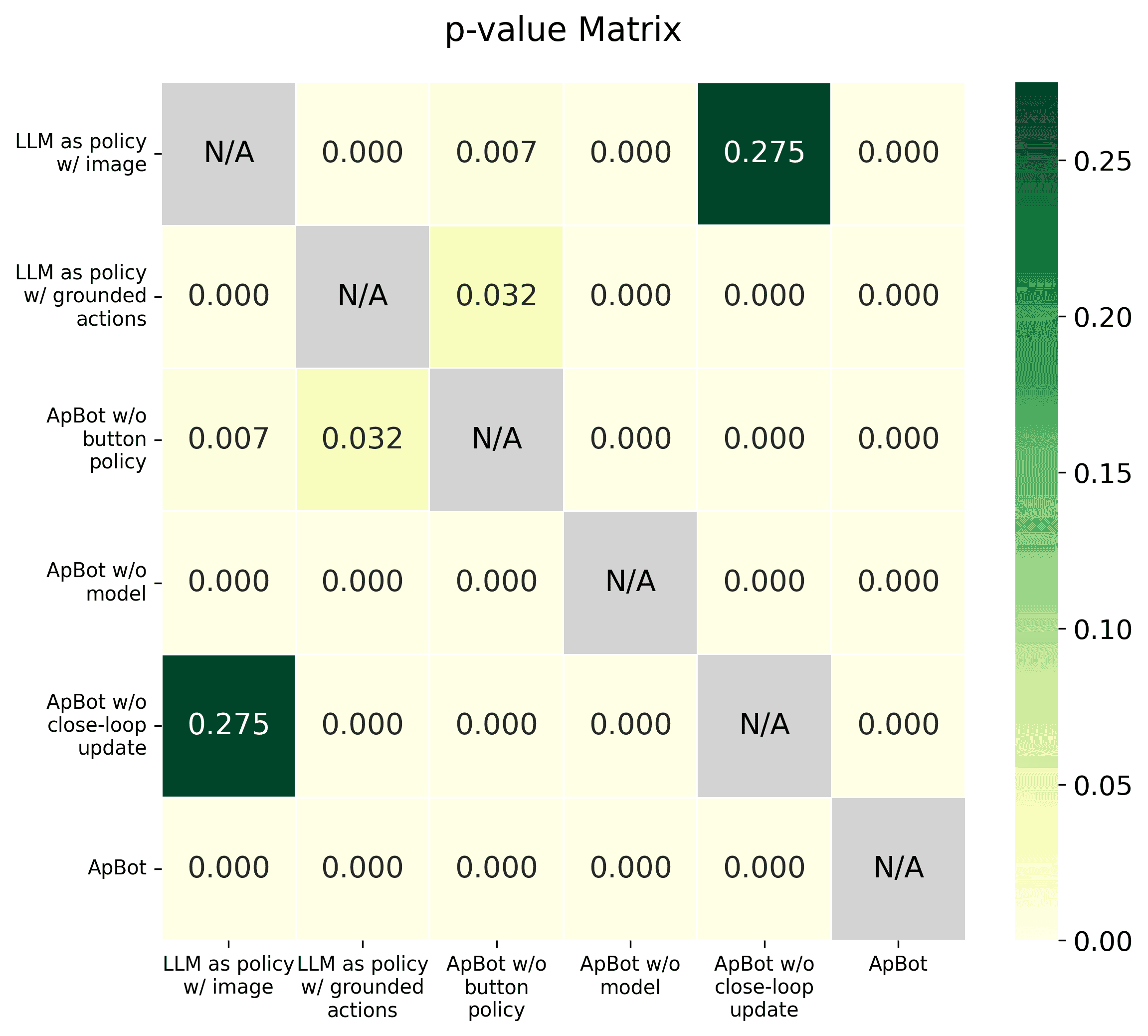}
  \caption{p-value matrix of all method pairs by $\chi^2$-test.}
  \vspace{-0.5cm}
  \label{fig:chi2test}
\end{wrapfigure}Detailed performance by appliance types is listed in \tabref{tab:overall_perform}. Referring to the rightmost figure in Fig. \ref{fig:overall_perform}, we can see that \myalgbr has shown robustness against appliance complexity (from left to right). \myalgbr does not suffer from severe performance drop when the number of involved variables increases. By contrast, both baselines suffer from significant performance degradation. We also conducted $\chi^2$ tests for all method pairs. As shown in Fig.~\ref{fig:chi2test}, performance differences are statistically significant for all pairs except \textit{\mynoupdate} and \textit{\cotimage}, indicating they are equally poor. Detailed analysis by appliance type is provided in \apref{ap:perform_details}. 

\reduceparagraphinterval
\paragraph{What are the main contributors of \myalg?}

We conduct ablations to evaluate the contributions of key components in \myalgbr. In summary, removing the structured appliance models (\textit{\mynomodel}) significantly degrades performance, mainly due to skipped steps or prematurely ending execution, which somehow mirrors the behavior of \textit{\cotaction}. This is because LVLMs cannot handle reasoning tasks involving a long history of many variables or constraints. It often ignores or hallucinates some of them (e.g., deciding whether the appliance is in the correct mode, proposing the required action to take), making the plan fail. Compared to \textit{\mynoreason}, we can conclude that invoking code to compute required action sequences (Sec.~\ref{sec:macro-action}) is crucial to ensure the correctness of generated policies. This is because LVLM agents struggle to assign variable values correctly when the variable range is large, when the transition $\trans$ is complex or when the variable value options are semantically similar. Finally, we find that closed-loop updates for home appliance models are critical. 
Performance of \textit{\mynoupdate} suffers a rapid, sharp drop as the complexity of appliances increases. It fails to recover from any model errors, like open-loop policies. It reveals that current state-of-the-art LVLMs still struggle with generating constrained structures correctly in one shot, like the models of home appliances. All these results illustrate the necessity of structured reasoning for robust appliance operation. We further provide qualitative examples in \apref{ap:perform_details} and failure analysis in \apref{ap:failures}.

\reduceparagraphinterval
\paragraph{What is the benefit of explicit action grounding? }

\begin{wrapfigure}{r}{0.4\textwidth}
    \vspace{0.1cm}
    \centering
    \includegraphics[width=0.38\textwidth]{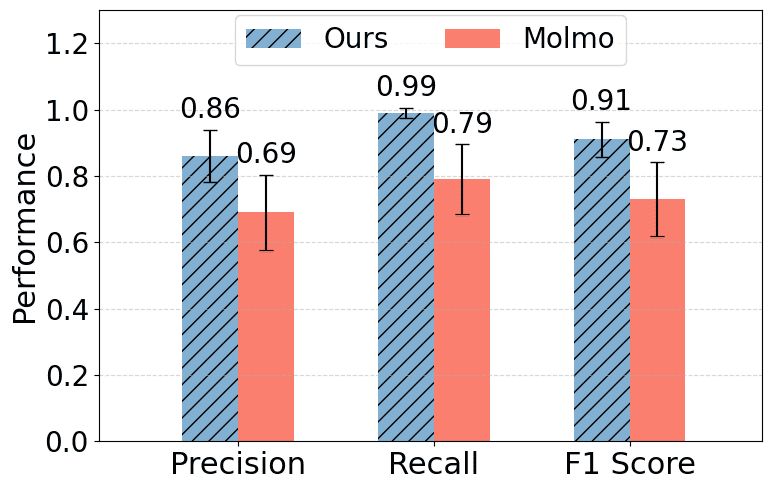}
    \caption{Comparison of action grounding performance between our method and Molmo. Standard deviation is across different appliance types.}
    \vspace{-0.5cm}
    \label{fig:act_ground}
\end{wrapfigure}

Our proposed method to ground actions can boost the overall performance of home appliance operation by 18\% on average by comparison with the performance between \textit{\cotimage} and \textit{\cotaction} in \figref{fig:overall_perform}, because LVLMs struggle with appliance button recognition.
To further investigate the effectiveness of our action grounding methods, we tested and provided the visual grounding results for symbolic actions based on control panel images.
The ground-truth labels of executable action regions are manually labeled. The comparison results between our method and \textit{Molmo} are shown in \figref{fig:act_ground}. Our method is statistically significantly better than \textit{Molmo} across all appliances (with $p$-value less than 0.001). 
The performance gain primarily comes from combining the advantages of (1) explicit text recognition, (2) high-recall detection, and (3) semantic understanding of graphical button icons of LVLMs. By contrast, \textit{Molmo} demonstrates reasonable text or symbol recognition ability, yet not robust enough as the specialist OCR models.

\reducesecinterval
\subsection{Deployment on Real-Robot} 

\reduceparagraphinterval

We deploy our method on a Kinova Gen3 arm and demonstrate its applicability to three household appliances: a blender, an induction cooker, and a water dispenser, each evaluated with semantically diverse instructions.
The button-pressing policy is parameterized by a bounding box. To compute the target end-effector pose, we compute the point cloud of the button and extract its surface normal. The robot aligns its gripper tip with the normal at a slightly tilted angle and moves 0.1 cm beyond the surface to ensure successful activation. More details can be founded in Appendix \ref{ap:realworld}.
\figref{fig:real_appliance_results} illustrates two example instructions carried out on the water dispenser and the induction cooker, each frame executing an action. With our method, the robot can reason about how to perform previously unseen, long-horizon operation tasks by referring to the user manual. Additional real-world demonstrations and results can be found in Appendix~\ref{ap:realworld} and the accompanying video.

\reducesecinterval
\section{Conclusion}
\reducesecinterval
We presented \myalg, a generalizable method that enables zero-shot operation of novel household appliances by referencing the user manual. 

By leveraging the structured model of appliances, \myalg demonstrates statistically significant robustness against diverse appliance types and language instructions. 

We built an evaluation suite including the benchmark of real-world appliances, manuals, open-ended task instructions, and symbolic simulators to benchmark home appliance operation. 

Results demonstrate that \myalg is able to improve the success rate significantly when compared with all baselines. 
We also deployed and demonstrated our system on real-world robotic tasks.
Results show that \myalg can reliably finish language-specified tasks with only the manual and visual observation as the inputs autonomously.

\section{Failure Analysis \& Limitations}

\paragraph{Failure Analysis.}
We analyzed 18 failed instructions and identified two major causes: action grounding errors (16.7\%) and modeling errors (83.3\%). 
Action grounding failures occur often when detectors misidentify soft-touch panels or icon-only buttons without physical boundaries. 
Modeling errors mainly arise from goal state ambiguity and consistent errors even after syntax checking. 
We demonstrate the details of failure cases in our experiments in \apref{ap:failures}.

\paragraph{Limitations.} 

First, \myalg does not support touchscreens, which are becoming prevailing in recent years. In the future, we would like to integrate the material design of the manipulator to support such kinds of interface.
Besides, button manipulation itself is a challenging task of robotics \cite{wang2018robot}. Currently, \myalg does not incorporate subtle modeling or feedback of diverse buttons (e.g., the frictions, tactile feedbacks), which are absolutely important for robust button manipulation. We will take advantage of tactile sensing in the future to support more robust manipulation \cite{yu2024octopi}.
Also, the action grounding module is not fully reliable, especially for buttons without clear physical boundaries or with icon-only symbols. In the future, we will try to specialize a button detector for robust button detection and grounding, or integrate a human-in-the-loop strategy \cite{xiao2024robi}.
Finally, \myalg does not consider complex manipulation skills for appliances, such as opening/closing doors, plugging, and putting in or removing items from the appliance container. We will leverage recent advances of manipulation policies in the future to support such kinds of sophisticated skills.

\bibliography{references}  

\appendix

\newpage
\section{Example of the Appliance Model and Simulation}
\label{ap:appliance_model}
\subsection{Structured Appliance Model Example}
Below is an example of the appliance model generated using \myalgbr for a \textit{dehumidifier}. 
It includes a list of variables extracted from the manual, the macro actions, and transitions.
During inference, we directly generate the model in the following format with the help of LVLM agents, based on which we further generate operation plans with Prompt \ref{prompt:generate_task_policy_and_goal_state}.
Note that the \textit{macro action}, which is typically a concept in computer science, is phrased as \textit{feature} in our prompts to match the commonly used term in most of the manuals.
In practice, we group consecutive actions of adjusting the same variable in a macro action into a \textit{step}, which empirically improves robustness and facilitates the syntax checking (\secref{ap:syntax_checker}).

\begin{tcolorbox}[colback=gray!5]
    \begin{minted}[breaklines, breakanywhere]{python}
# Variables of the appliance defined in the State
variable_power_on_off = DiscreteVariable(value_range=["on", "off"], current_value="off")
variable_fan_speed = DiscreteVariable(value_range=["low", "mid", "high"], current_value="low")
...

# Macro actions
feature_list = {}

feature_list["turn_on_off"] = [
    {"step": 1, "actions": ["press_power_button"], "variable": "variable_power_on_off", "step_size": 2}
]
feature_list["adjust_fan_speed"] = [
    {"step": 1, "actions": ["press_speed_button"], "variable": "variable_fan_speed", "step_size": 3}
]
...

# Transitions
simulator_feature = Feature(feature_list=feature_list, current_value=("empty", 1))

class Simulator(Appliance):

    def reset(self):
        self.feature = simulator_feature
        self.variable_power_on_off = variable_power_on_off
        self.variable_fan_speed = variable_fan_speed
        ...

    def press_power_button(self):
        self.feature.update_progress("press_power_button")
        self.execute_action_and_set_next("press_power_button")

    def press_speed_button(self):
        self.feature.update_progress("press_speed_button")
        self.execute_action_and_set_next("press_speed_button")

    ...

    \end{minted}
\end{tcolorbox}

\newpage

\section{Details of the Syntax Checker}
\label{ap:syntax_checker}

To mitigate hallucination during appliance model generation, we implement a suite of syntax checkers to further validate the generated models, mainly for the macro actions, transitions, and goal specifications. 
Additionally, all generated codes are verified to ensure that they fit the required output format, with the help of \textit{regular expressions}. The detailed prompt can be found in Prompt~\ref{prompt:extract_features}. We list the syntax checkers here:

\begin{enumerate}[label=\arabic*.]

\item \textbf{Missing Variable:}  

Every step should adjust some variables.

\item \textbf{Empty or Non-Existent Action:}  
Each step should contain at least one valid action.  

\item \textbf{Action Coverage:}  
Every action in $\aacts$ should appear in some macro actions.  

\item \textbf{Variable Coverage.}  
Every variable defined in the state space $\astates$ should appear in some macro actions.

\item \textbf{Duplicate Action Sequences:} 
We check if there are possibly duplicate action sequences (e.g., set a variable to a specified value twice).

\item \textbf{Number-Pad Action Compatibility:}  
Number-pad actions should not appear when modeling appliances without a number pad.

\item \textbf{Input String Reset:}  
The appliance with a number pad should reset the input string of the number pad whenever it switches away.

\item \textbf{Action-Variable Consistency:} Actions should only adjust associated variables.  

\item \textbf{Goal Validity:}  
$\agstates$ should be fully specified, i.e., each variable should be assigned or intentionally ignored.

\end{enumerate}

\section{Details of State Estimation and Model Updates}

\label{ap:model_updates}

\paragraph{State Estimation.} 
The robot estimates the appliance state using two feedback modalities. In simulation, textual feedback directly provides ground truth values for state variables being tuned. In real-world scenarios, the robot captures an image of the appliance and uses LVLM agents to convert visual observations into textual state descriptions. After completing each macro action, the robot compares the predicted state resulting from the planned action with the observed state extracted from feedback. The result indicates whether the macro action successfully achieved its intended effect.

\paragraph{Model Updates.} 
The generated operation plan is executed in the minimal unit of a macro action. The robot does not track states or update appliance models during the execution of a macro action, but only after its completion. 
If the observed state does not match the predicted state, it indicates that some transitions in this macro action might be wrong.
Hence, the robot initiates a sequence of exploration actions to explore and fix the possible errors.
Empirically, we found that go-to transitions in $\atransgo$ are mostly correct.
Therefore, for each action in $\aactnb$, the robot continuously executes it until a previously observed state is observed again.
This exploration strategy is based on the observation that most actions in $\aactnb$ for appliances are circular.
Based on the observed state transitions, the robot updates the transition model regarding the corresponding action and regenerates all macro actions that depend on it.

\paragraph{An Example of State Estimation and Model Updates.}

Below is an example illustrating how the appliance model is automatically updated using closed-loop feedback.
Consider a task that requires turning on the fan and setting the fan speed to \texttt{high}. The robot begins by executing the \texttt{press\_power\_button} action under the macro action of \texttt{turn\_on\_off}. Upon receiving feedback indicating \texttt{power = on}, it invokes Prompt~\ref{prompt:compare_goal_state_with_feedback} to confirm that the subgoal is achieved as expected, i.e., the prediction matches the observation. 
Next, it proceeds to the macro action of \texttt{adjust\_fan\_speed}. Assuming the current speed is \texttt{low}, the robot executes the action sequence in this macro action.
Given a ground-truth cyclic variable range of \{\texttt{low}, \texttt{medium}, \texttt{high}\}, assume that the current action sequence is wrong, for example, requiring 1 press of the speed button, which results in \texttt{medium} speed.
After executing the planned actions, the feedback again indicates \texttt{speed = medium}, suggesting the goal is not yet met. The robot continues pressing the speed button and observes the feedback sequence: \texttt{medium} $\rightarrow$ \texttt{high} $\rightarrow$ \texttt{low} $\rightarrow$ \texttt{medium}. The repeated value \texttt{medium} indicates the entire value range has been cycled.
Using Prompt~\ref{prompt:diagnose_incorrect_variable_definition}, the robot diagnoses that the transition of the action \texttt{adjust\_fan\_speed} was incorrect. It then updates the model according to the diagnosis. 
In detail, it sequentially updates the variable definition via Prompt~\ref{prompt:update_variable_definition}, revises the appliance model using Prompt~\ref{prompt:update_appliance_model}, and adjusts the goal state accordingly using Prompt~\ref{prompt:update_goal}.
With the updated current value of \texttt{medium}, the robot re-plans the action sequence and presses the speed button once more. This time, the feedback confirms \texttt{speed = high}, and Prompt~\ref{prompt:compare_goal_state_with_feedback} verifies that the goal is satisfied. The task completes successfully.

\section{Details of the Evaluation Benchmark}
\label{ap:eval_dataset}

\subsection{Appliances Categories and Data Collection}
\label{ap:appliances}

\begin{figure}[t]
    \centering
    \includegraphics[width=1.0\linewidth]{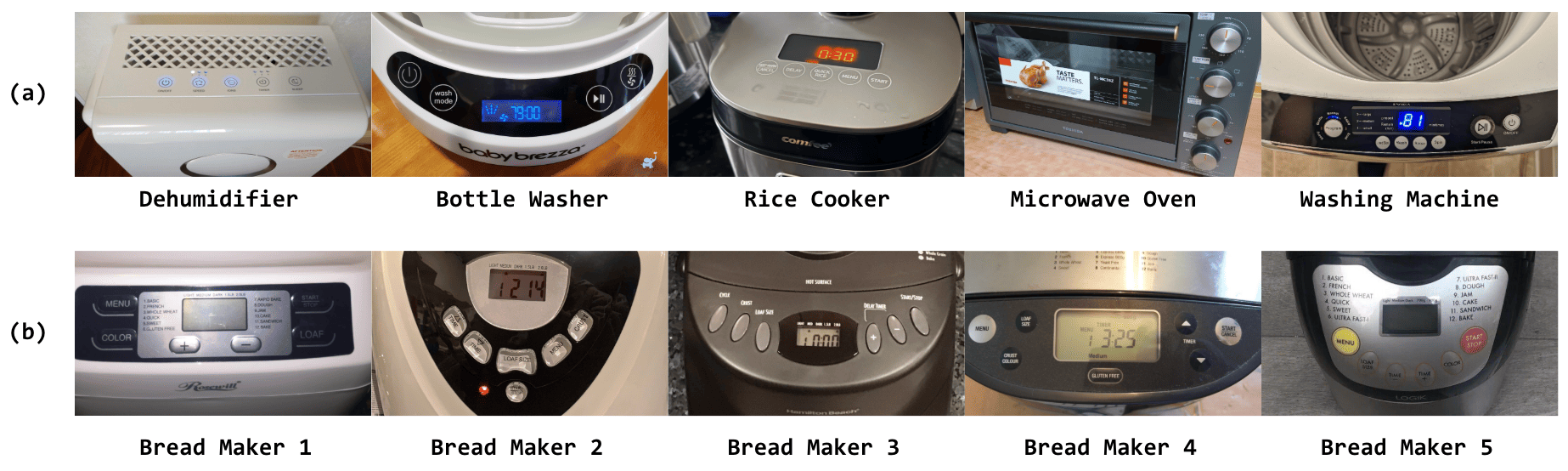}
    \caption{Appliances in our benchmark. (a) Appliance Types. (b) All Instances of Bread Maker.}
    \label{fig:sample_panel}
\end{figure}

\begin{figure}[t]
    \centering
    \includegraphics[width=1.0\linewidth]{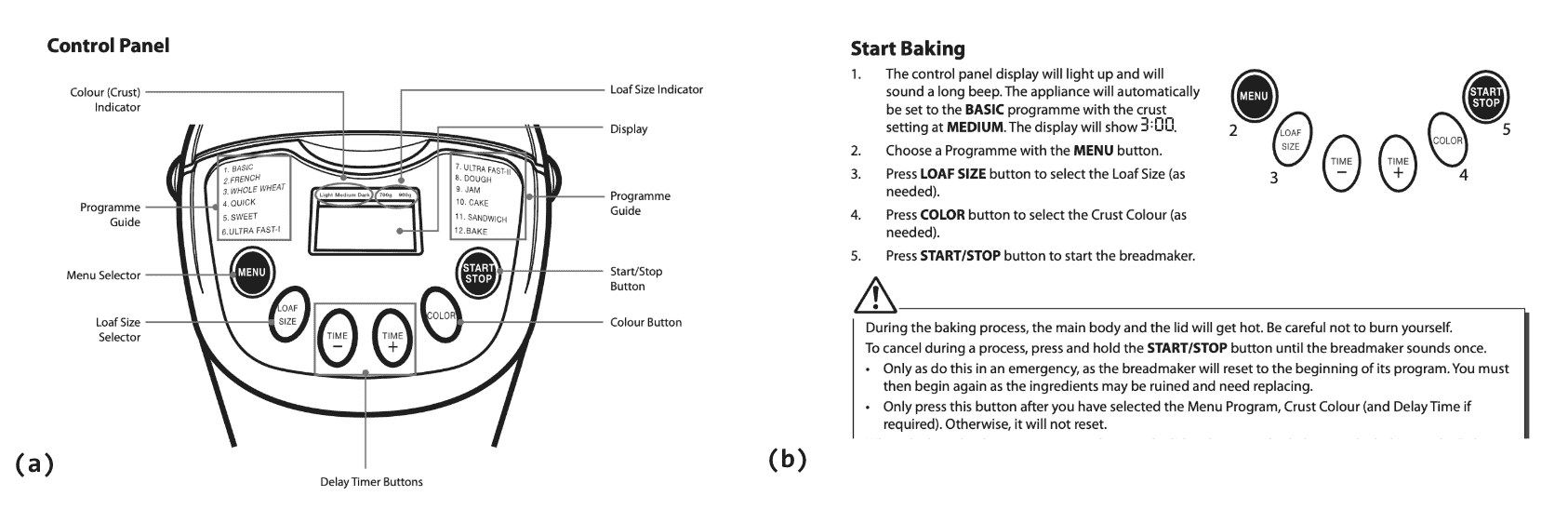}
    \caption{An example user manual for the Bread Maker. (a) Control panel. (b) Unstructured step-by-step procedure for a macro action.}
    \label{fig:sample_manual}
\end{figure}

The benchmark covers six types of household appliances: \textit{dehumidifiers}, \textit{bottle washers}, \textit{rice cookers}, \textit{microwave ovens}, \textit{bread makers}, and \textit{washing machines}. These categories were chosen for their differences in mechanism complexity and functional diversity. As shown in Figure~\ref{fig:sample_panel}, each category includes five distinct instances, resulting in 30 appliances in total.

For each instance, we collect an image of the control panel from the Internet, including Amazon and eBay. We also collect the corresponding user manual from the product’s official website or support page. From each manual, we extract two key parts:  
(1) a control panel legend that links interface elements to their locations (see ~\figref{fig:sample_manual}a), and  
(2) step-by-step instructions that describe how to operate specific features (see ~\figref{fig:sample_manual}b).
They are used to construct the structured symbolic model for each appliance. The appliances vary in interface layout and the number of adjustable variables. To ensure fair comparison, we assign a fixed number of target variables per appliance type, subject to their inherent complexity. For example, \textit{dehumidifiers} require adjusting one variable, while washing machines require six. 

\subsection{Task Instructions}
\label{ap:tasks}

We design 300 goal-directed natural language instructions, with 10 tasks per appliance instance. Each instruction specifies a clear goal by assigning specific target values to a set of variables. 
This ensures consistency across methods and focuses evaluation on symbolic reasoning and execution.
Ground-truth values are manually labeled to support automatic evaluation.
The number of variables involved in each task depends on the type of appliance, facilitating controllable comparison. For example, all instructions for \textit{dehumidifier} involve only one variable. More complex ones, like washing machines, involve up to six variables. Details and samples of instructions are listed in \tabref{tab:instruction_complexity}. 

\begin{table}[t]
\centering
\caption{Examples of Task Instructions for Different Appliances}
\vspace{0.5em}
\renewcommand{\arraystretch}{2}
\resizebox{\textwidth}{!}{%
\begin{tabular}{>{\centering\arraybackslash}m{1cm}>{\centering\arraybackslash}m{2.5cm}>{\centering\arraybackslash}m{8.2cm}>{\centering\arraybackslash}m{7cm}}
\toprule
\textbf{\# Vars} & \textbf{Appliance} & \textbf{Sample Instruction} & \textbf{Target Settings} \\
\midrule
\parbox[c][2em][c]{1cm}{\centering 1} & \parbox[c][2em][c]{2.5cm}{\centering Dehumidifier} & 
\parbox[c][2em][c]{8.2cm}{\raggedright “Set the humidity to 50\%.”} &
\parbox[c][2em][c]{7cm}{\begin{itemize}[leftmargin=*, itemsep=0pt]
\item Humidity = 50\%
\end{itemize}} \\
\hline
\parbox[c][3.2em][c]{1cm}{\centering 2} & \parbox[c][3.0em][c]{2.5cm}{\centering Bottle\\Washer} & 
\parbox[c][3.2em][c]{8.2cm}{\raggedright “Power on the device and initiate a 45-minute automatic sterilization and drying cycle.”} &
\parbox[c][3.2em][c]{7cm}{%
\begin{minipage}[t]{0.49\linewidth}
\begin{itemize}[leftmargin=*, itemsep=0pt]
\item Power = On
\end{itemize}
\end{minipage}%
\begin{minipage}[t]{0.49\linewidth}
\begin{itemize}[leftmargin=*, itemsep=0pt]
\item Drying Time = 45 min
\end{itemize}
\end{minipage}} \\
\hline
\parbox[c][3.5em][c]{1cm}{\centering 3} & \parbox[c][3.0em][c]{2.5cm}{\centering Rice\\Cooker} & 
\parbox[c][3.5em][c]{8.2cm}{\raggedright “Adjust the delay timer to 30 minutes, set the rice cooker to White Rice mode, and start the operation.”} &
\parbox[c][3.5em][c]{7cm}{%
\begin{minipage}[t]{0.49\linewidth}
\begin{itemize}[leftmargin=*, itemsep=0pt]
\item Menu = White Rice
\item Start = On
\end{itemize}
\end{minipage}%
\begin{minipage}[t]{0.49\linewidth}
\begin{itemize}[leftmargin=*, itemsep=0pt]
\item Delay Timer = 30 min
\end{itemize}
\end{minipage}} \\
\hline
\parbox[c][5.5em][c]{1cm}{\centering 4} & \parbox[c][5.5em][c]{2.5cm}{\centering Microwave\\Oven} & 
\parbox[c][5.5em][c]{8.2cm}{\raggedright “Set the upper tube temperature to 150°C. Select the cooking function as 'upper and lower heating tube'. Then set the lower tube temperature to 150°C and adjust the cooking time to 20 minutes.”} &
\parbox[c][5.5em][c]{7cm}{%
\begin{minipage}[t]{0.49\linewidth}
\begin{itemize}[leftmargin=*, itemsep=0pt]
\item Upper Temp = 150°C
\item Lower Temp = 150°C

\end{itemize}
\end{minipage}%
\begin{minipage}[t]{0.49\linewidth}
\begin{itemize}[leftmargin=*, itemsep=0pt]
\item Time = 20 min

\end{itemize}
\end{minipage}

\vspace{0.3em} 
\begin{minipage}[t]{0.91\linewidth}
\begin{itemize}[leftmargin=*, itemsep=0pt]
\item  Function = Upper / Lower Heating

\end{itemize}
\end{minipage}}
\\
\hline
\parbox[c][4.5em][c]{1cm}{\centering 5} & \parbox[c][4.5em][c]{2.5cm}{\centering Bread\\Maker} & 
\parbox[c][4.5em][c]{8.2cm}{\raggedright “Bake a large, medium-crust French loaf using the French menu. Set a 2-hour delay timer, then start the bread maker.”} &
\parbox[c][4.5em][c]{7cm}{%
\begin{minipage}[t]{0.49\linewidth}
\begin{itemize}[leftmargin=*, itemsep=0pt]
\item Menu = French
\item Crust = Medium
\item Start = On
\end{itemize}
\end{minipage}%
\begin{minipage}[t]{0.49\linewidth}
\begin{itemize}[leftmargin=*, itemsep=0pt]
\item Loaf Size = Large
\item Delay = 2 hrs
\end{itemize}
\end{minipage}} \\
\hline
\parbox[c][5em][c]{1cm}{\centering 6} & \parbox[c][5em][c]{2.5cm}{\centering Washing\\Machine} & 
\parbox[c][5em][c]{8.2cm}{\raggedright “Turn on the washing machine. Select the Normal program for everyday clothes, set the water level to 55 L, schedule it to finish in 4 hours, start the machine, and activate the child lock.”} &
\parbox[c][5em][c]{7cm}{%
\begin{minipage}[t]{0.49\linewidth}
\begin{itemize}[leftmargin=*, itemsep=0pt]
\item Power = On
\item Water Level = 55 L
\item Start = On
\end{itemize}
\end{minipage}%
\begin{minipage}[t]{0.49\linewidth}
\begin{itemize}[leftmargin=*, itemsep=0pt]
\item Program = Normal
\item Preset = 4 hrs
\item Child Lock = On
\end{itemize}
\end{minipage}} \\
\bottomrule
\end{tabular}%
}
\label{tab:instruction_complexity}
\end{table}

\subsection{Simulators and Ground Truth Feedback}
\label{ap:simulator}
Each appliance instance is paired with a symbolic simulator implemented in Python, providing a deterministic testing environment. The simulator encodes adjustable variables and valid actions based on the appliance manual, preserving constraints defined by its manual. It also defines executable regions tied to control panel elements.  
Actions are input as a pair: a bounding box and an action type (e.g., press or turn). If the action is valid, the simulator updates the variable state and returns a textual message (e.g., \texttt{temperature = 150\textdegree C}) indicating the resulting variable value. 

\subsection{Metrics}

We mainly evaluate \textit{Success Rate}, \textit{Average Step}, and \textit{Success weighted by Path Length}. Besides, we also report the \textit{Execution Step} in \apref{ap:perform_details}. 

\paragraph{Success Rate:} It is defined as the proportion of tasks completed successfully before exceeding 25 \textit{reasoning steps}. The number of reasoning steps is defined as the total number of macro actions, excluding the exploration steps. We define \textit{success} as the achievement of variable values included in the specified goal state at the end of execution.

\paragraph{Average Step:} This metric indicates the average number of \textit{reasoning steps}, i.e., the number of macro actions excluding the exploration, taken before either achieving success or reaching the maximum number (25 in our experiments). This metric focuses on the reasoning efficiency of the system.

\paragraph{Success weighted by Path Length (SPL):} SPL evaluates success while considering the actual number of physical \textit{execution steps}, i.e., symbolic actions in $\aacts$. The optimal number of actions is manually labeled by a human oracle. Intuitively, this metric evaluates the efficiency considering both execution and exploration.

\paragraph{Execution Step:} It indicates the average number of symbolic actions in $\aacts$, i.e., the number of actions that the robot actually executes physically, including the exploration ones, taken before success or reaching the max reasoning steps. It evaluates the physical execution efficiency of algorithms.

\section{Details of Experimental Settings}

\label{ap:experiment-setup}
\subsection{Detailed Settings of LVLMs}
\label{ap:model_setting}
Table~\ref{tab:lvml_hyperparams} lists non-default hyperparameters of all models used in our experiments. GPT-4o was used for both appliance model construction and action grounding. Claude-3.5, EasyOCR, and OWLv2 were used only for action grounding. In real-world experiments, where control panels are simpler, only OWLv2 was used for control element detection to improve efficiency; EasyOCR and SAM were omitted.

\begin{table}[t]
\centering
\sloppy
\caption{Hyperparameters of Used Models.}
\label{tab:lvml_hyperparams}
\begin{tabularx}{\textwidth}{>{\bfseries}p{2cm} p{2.5cm} X p{2.5cm}}
\toprule
Component & Parameter Name & Explanation & Value \\
\midrule
\multirow{3}{*}{\textbf{GPT}} 
& \texttt{model} & GPT model version. & \texttt{\makecell[l]{GPT-4o-\\2024-11-20}} \\
& \texttt{temperature} & Controls output randomness. & \texttt{1.0} \\
& \texttt{top\_p} & Nucleus sampling cutoff. & \texttt{1} \\
\midrule

\multirow{2}{*}{\textbf{OWLv2}} & \texttt{model} & OWLv2 model name. & \texttt{\makecell[l]{owlv2-large-\\patch14-ensemble}} \\
& \texttt{box\_threshold} & Object detection threshold. & \texttt{0.5} \\
\midrule
\multirow{3}{*}{\textbf{EasyOCR}} 
& \texttt{text\_threshold} & Text detection threshold. & \texttt{0.5} \\
& \texttt{low\_text} & Includes blurry text. & \texttt{0.4} \\
& \texttt{contrast\_ths} & Contrast enhancement threshold. & \texttt{0.05} \\
\midrule
\multirow{2}{*}{\makecell{\textbf{Segment}\\\textbf{Anything}\\\textbf{Model}}} 
& \texttt{iou} & IoU threshold for masks. & \texttt{0.1} \\
& \texttt{conf} & Mask confidence filter. & \texttt{0.9} \\
\bottomrule
\end{tabularx}
\end{table}

\subsection{Baselines}
\label{ap:baselines}
Table~\ref{tab:component_ablation} summarizes the key components in each method. \tick indicates the component is present in the corresponding method; {\cross}\ indicates it is omitted or replaced directly by LVLM equivalents. The prompts used for all methods can be found in Appendix~\ref{ap:prompts}.

\paragraph{Grounded Action} refers to whether the method reasons over symbolic actions, which are visually grounded to control panel elements via our action grounding method (see \secref{sec:action-grounding}), or directly reasons over image regions without symbolic abstraction.

\paragraph{Model} $\overline{\mathcal{M}}$ indicates whether the method constructs and follows a symbolic appliance model extracted from the user manual. If present, variable adjustments follow a fixed sequence specified by macro actions $\amacroaction$ and the task policy $\pi$, instead of being chosen reactively by an LLM.

\paragraph{Button Policy} denotes whether action sequences for variable adjustment are computed using predefined transition functions $\atrans$, rather than being generated by LLMs.

\paragraph{Closed-loop Update} refers to whether the method incorporates execution feedback to update state estimation and re-generate action sequences. Methods without this component operate in an open-loop manner, executing fixed sequences or reasoning reactively without correcting for execution errors.

\begin{table}[H]
\centering
\caption{Baseline methods and ablation of key components in \myalgbr. \tick indicates the component is used.}
\label{tab:component_ablation}
\begin{tabularx}{\textwidth}{>{\raggedright\arraybackslash}p{4.5cm} *{4}{>{\centering\arraybackslash}X}}
\toprule
\textbf{Method} & \textbf{Grounded} & \textbf{Model} $\overline{\mathcal{M}}$ & \textbf{Button} & \textbf{Closed-loop} \\
& \textbf{Action} &  & \textbf{Policy} & \textbf{Update} \\
\midrule
\cotimage     & \cross & \cross & \cross & \tick \\
\cotaction    & \tick  & \cross & \cross & \tick \\
\cmidrule(lr){1-5}
\mynomodel    & \tick  & \cross & \tick  & \tick \\
\mynoreason   & \tick  & \tick  & \cross & \tick \\
\mynoupdate   & \tick  & \tick  & \tick  & \cross \\
\myalg        & \tick  & \tick  & \tick  & \tick \\
\bottomrule
\end{tabularx}
\end{table}

\newpage
\section{Details of Performance}
\label{ap:perform_details}

\begin{figure}[h]
     \centering
     \includegraphics[width=1.0\linewidth]{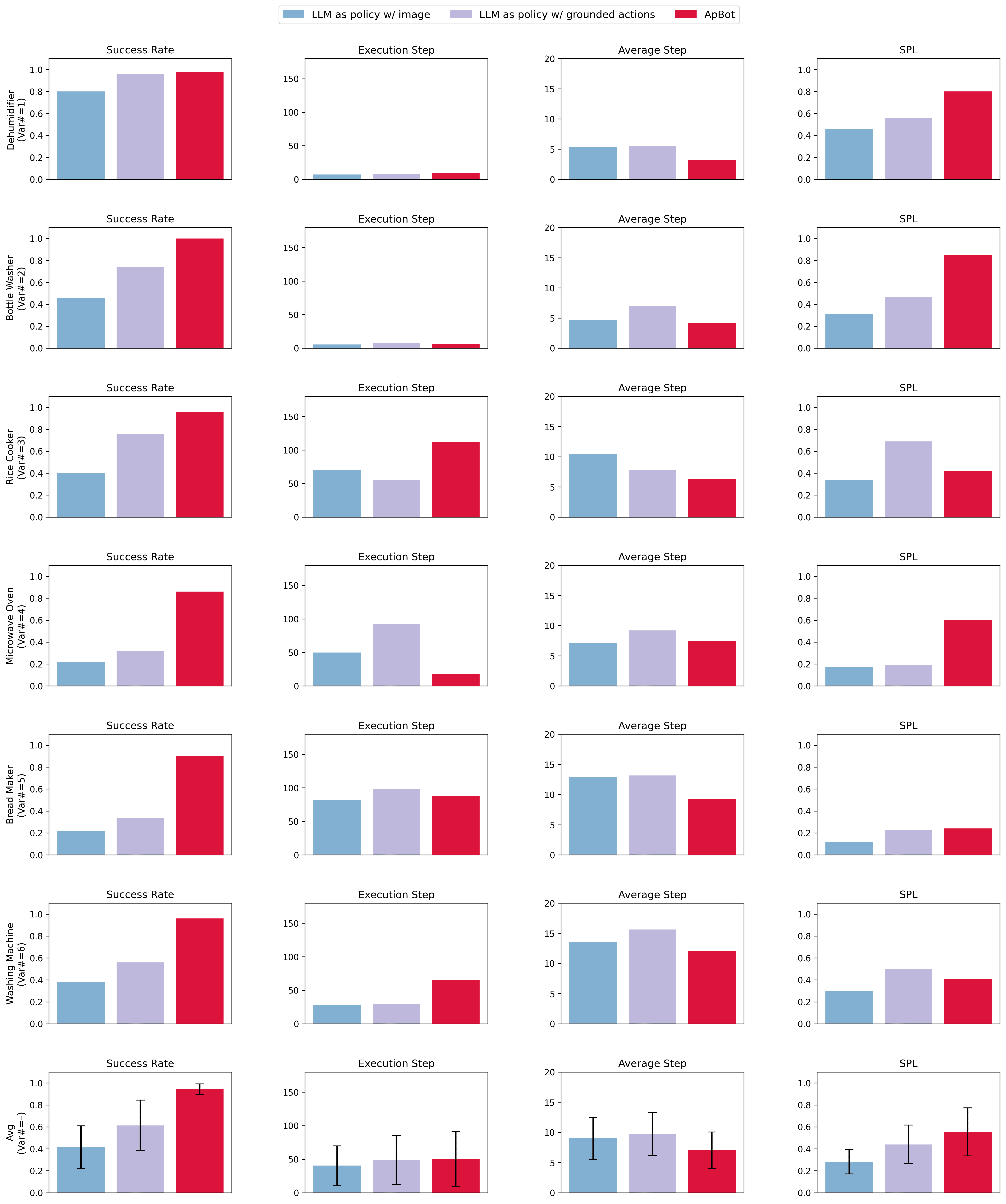}
     \vspace{-10pt}
     \caption{Performance of home appliance operation by appliance type, including average task success rate (SR), average number of execution steps (Average Steps), and SPL (Success weighted by Path Length) on baseline methods.}
     \label{appendix:accuracy_by_type_baseline}
     \vspace{-15pt}
\end{figure}

\begin{figure}[h]
     \centering
     \includegraphics[width=1.0\linewidth]{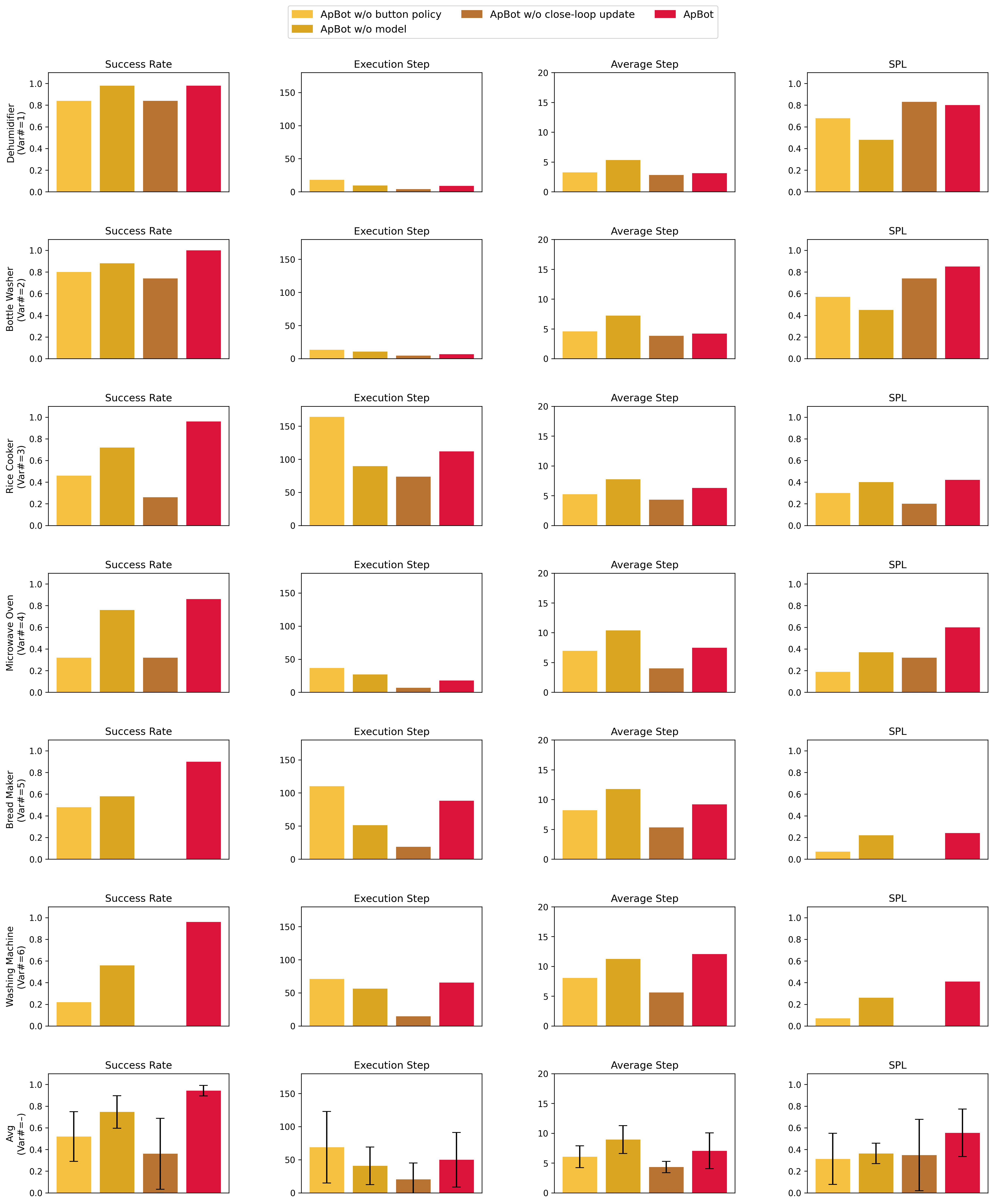}
     \vspace{-10pt}
     \caption{Performance of home appliance operation by appliance type, including average task success rate (SR), average number of execution steps (Average Steps), and SPL (Success weighted by Path Length) on ablation methods.}
     \label{appendix:accuracy_by_type_ablation}
     \vspace{-15pt}
\end{figure}

\begin{figure}[h]
     \centering
     \includegraphics[width=0.7\linewidth]{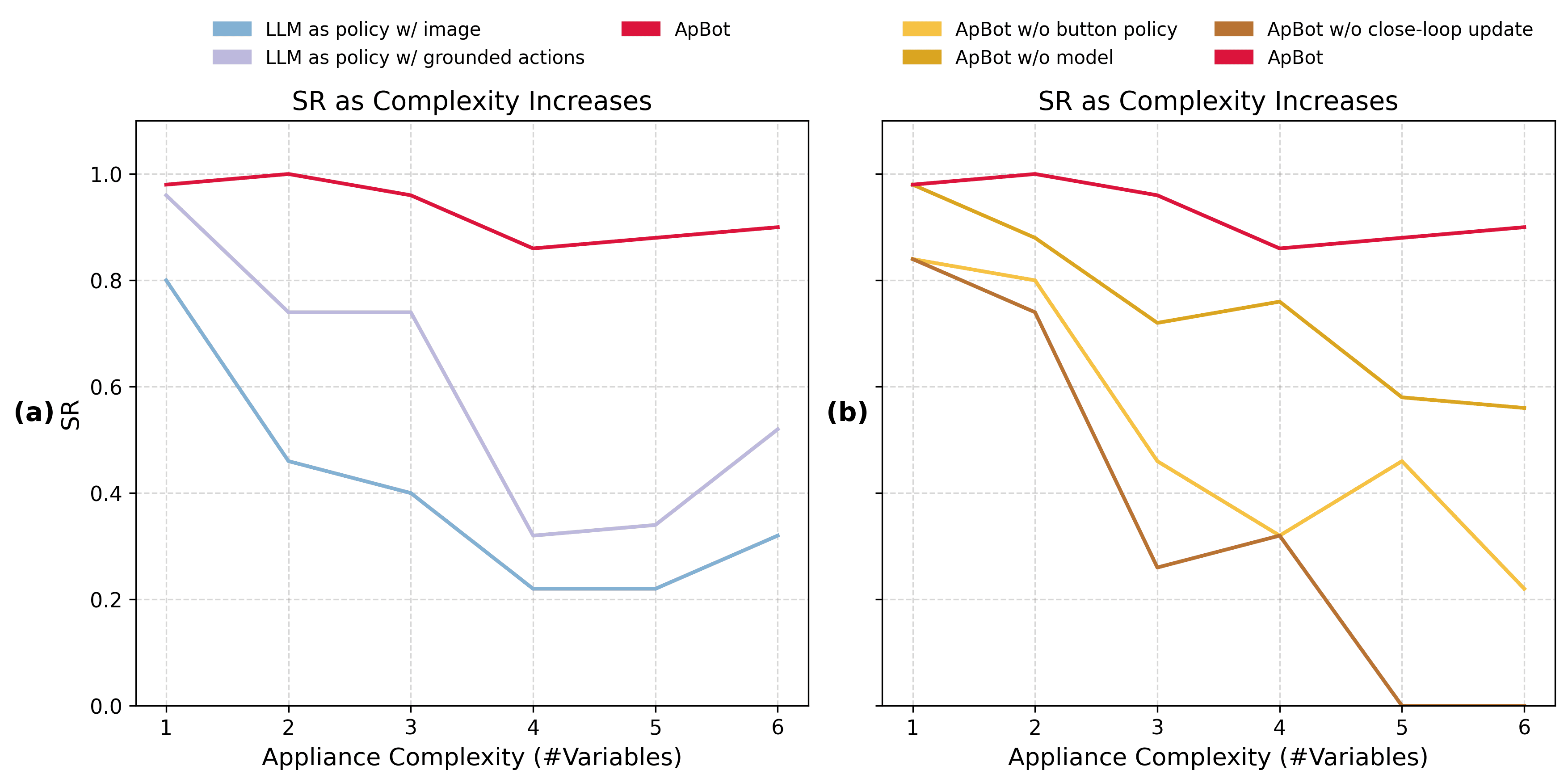}
     \caption{Average task success rate (SR) by increasing variable size conditioned on appliance type.}
     \label{fig:sr_vs_complexity}
     
\end{figure}

Figure~\ref{appendix:accuracy_by_type_baseline} and Figure~\ref{appendix:accuracy_by_type_ablation} shows the performance of \myalgbr on six appliance types. Each appliance type has a different number of variables to adjust, from 1 to 6 (top to bottom). As the number of variables increases, \myalg does not suffer a severe performance drop in terms of success rate (Figure~\ref{fig:sr_vs_complexity}). By contrast, baseline methods like \textit{\cotimage} and \textit{\cotaction} drop significantly on tasks with more variables. This shows that structured models, structured reasoning, and closed-loop updates help in handling complex tasks. Another interesting observation is that SPL suffers from an obvious drop when increasing the complexity of appliances. The reason is that for complex appliances and tasks, there will always be more modeling errors, which require more exploration steps for model updates. This further demonstrates the necessity of appliance modeling and online updates. 

 \begin{figure}[t]
     \centering
     \includegraphics[width=1.0\linewidth]{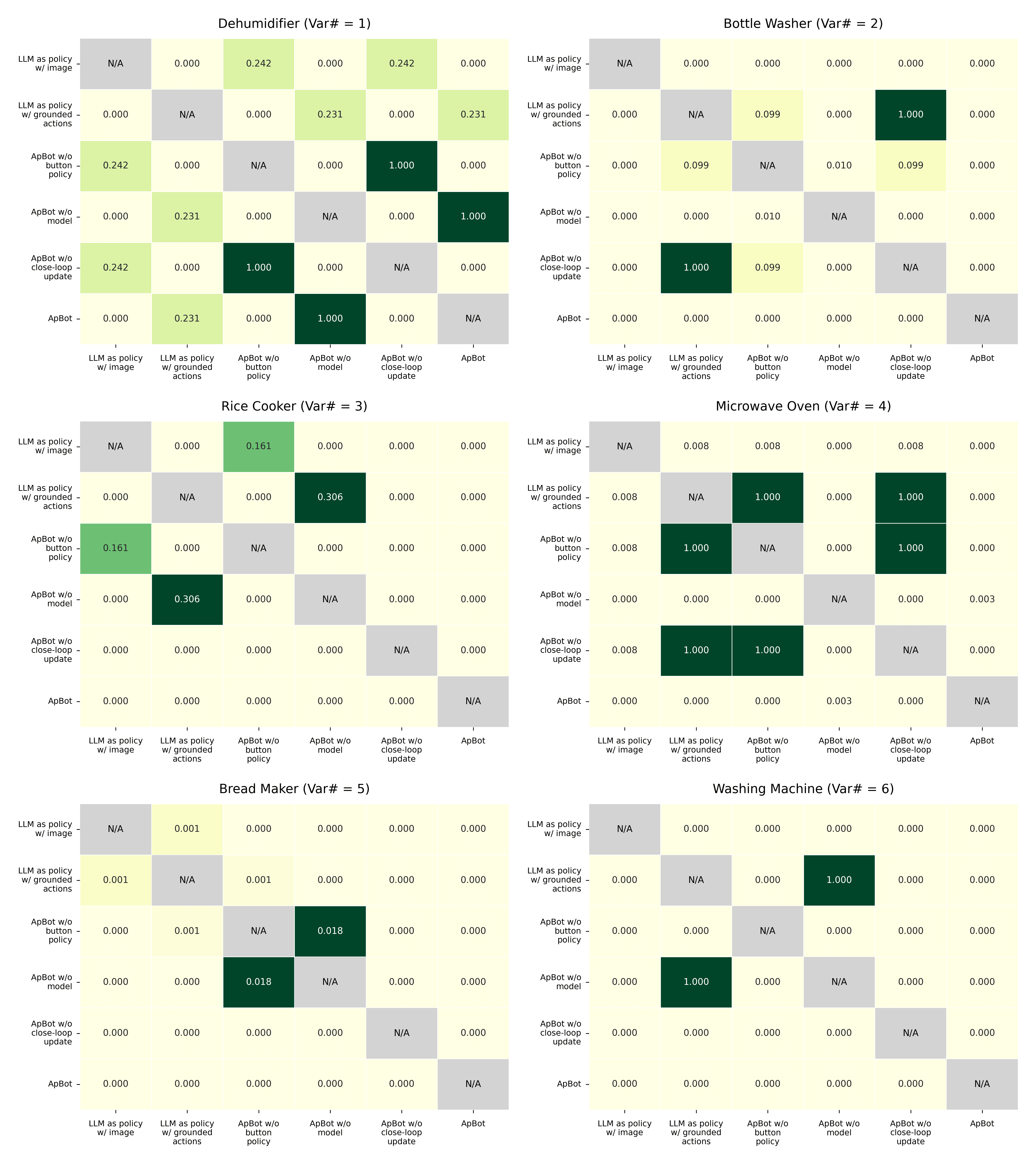}
     \vspace{-10pt}
     \caption{p-value matrix of all method pairs by  $\chi^2$-test on different appliance types.}
     \label{appendix:p_value_by_type}
     \vspace{-15pt}
 \end{figure}

Figure~\ref{appendix:p_value_by_type} presents pairwise $\chi^2$-test p-values across six methods for each appliance type, with diagonal entries marked as "N/A". Each subplot corresponds to an appliance type, ordered by the number of variables to adjust per user instruction. As the number of variables increases, the performance gap between \myalgbr and baseline methods such as \textit{\cotimage} and \textit{\cotaction} becomes more statistically significant.

\begin{figure}[t]
     \centering
     \includegraphics[width=1.0\linewidth]{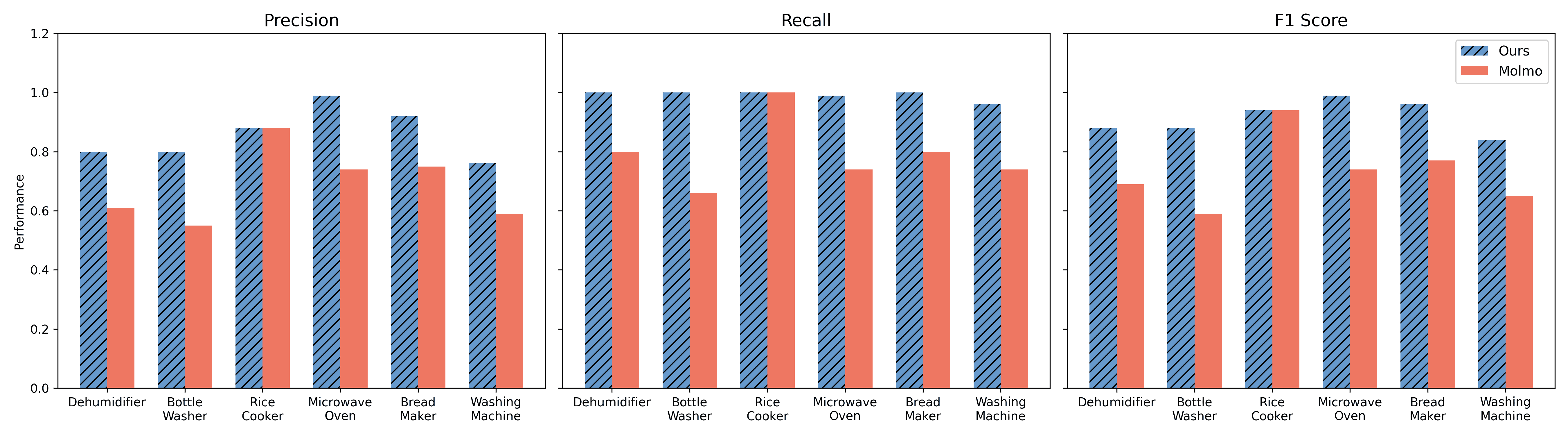}
     \vspace{-10pt}
     \caption{Comparison of action grounding performance between our method and Molmo on precision and recall across appliance types.}
     \label{appendix:grounding}
     \vspace{-15pt}
 \end{figure}

Figure~\ref{appendix:grounding} compares action grounding performance between \myalgbr and Molmo across six appliance types, evaluated by precision, recall, and F1 score. \myalgbr consistently outperforms Molmo on all appliance types, particularly on appliances with symbolic, iconic, or multi-word text labels, where a structured grounding procedure performs better.

\begin{table}[t]
\centering
\caption{Detailed Performance of Success Rate / Average Steps by Appliance Types}
\vspace{0.5em}
\resizebox{\textwidth}{!}{%
\begin{tabular}{lccccccc}
\toprule
\textbf{Method} & \makecell{\textbf{Dehumid-} \\ \textbf{ifier}} & \makecell{\textbf{Bottle} \\ \textbf{washer}} & \makecell{\textbf{Rice} \\ \textbf{cooker}} & \makecell{\textbf{Microwave} \\ \textbf{oven}} & \makecell{\textbf{Bread} \\ \textbf{maker}} & \makecell{\textbf{Washing} \\ \textbf{machine}}\\
\midrule
\cotimage & 0.80 / 5.34 & 0.46 / 4.64 & 0.40 / 10.46 & 0.22 / 7.12 & 0.22 / 12.88 & 0.32 / 13.46  \\
\cotaction & 0.96 / 5.48 & 0.74 / 6.94 & 0.74 / 7.88 & 0.32 / 9.22 & 0.34 / 13.02 & 0.52 / 15.36  \\
\mynoreason & 0.84 / 3.28 & 0.80 / 4.58 & 0.46 / 5.22 & 0.32 / 6.98 & 0.46 / 8.18 & 0.22 / 8.04  \\
\mynomodel & 0.98 / 5.32 & 0.88 / 7.22 & 0.72 / 7.80 & 0.76 / 10.38 & 0.58 / 11.92 & 0.56 / 11.44  \\
\mynoupdate & 0.84 / 2.82 & 0.74 / 3.82 & 0.26 / 4.34 & 0.32 / 4.00 & 0.00 / 5.34 & 0.00 / 5.62  \\
Ours & 0.98 / 3.12 & 1.00 / 4.22 & 0.96 / 6.28 & 0.86 / 7.46 & 0.88 / 9.14 & 0.90 / 11.64  \\
\bottomrule
\end{tabular}%
}
\label{tab:overall_perform}
\end{table}

We illustrate an online model update example triggered by a transition failure. An incorrect transition rule for the microwave function dial extracted from the user manual led to a goal mismatch. Upon observing inconsistent state feedback, \myalgbr exhaustively explores the function dial’s state space and updates the macro action to reflect the correct transition mapping.

\begin{tcolorbox}[title=Update Macro Action: Adjust Microwave Function, breakable, colback=gray!5]

\textbf{Action Applied:} (`turn\_function\_dial\_clockwise', 1)

\vspace{0.5em}
\textbf{Feedback Received:}
\begin{itemize}
    \item Fermentation
\end{itemize}

\vspace{0.5em}
\textbf{Goal Comparison:}
\begin{itemize}
    \item \textbf{Expected:} Lower \& Upper Heater
    \item \textbf{Observed:} Fermentation
    \item \textbf{Result:} Values are semantically different. Goal not reached.
\end{itemize}

\vspace{0.5em}
\textbf{Execution Trace:}

\begin{tabularx}{\textwidth}{p{5.2cm} X}
\toprule
\textbf{Action} & \textbf{Observed Value} \\
\midrule
(`turn\_function\_dial\_clockwise', 1) & Fermentation \\
(`turn\_function\_dial\_clockwise', 1) & Lower heater \\
(`turn\_function\_dial\_clockwise', 1) & Upper heater \\
(`turn\_function\_dial\_clockwise', 1) & Lower \& upper heater \\
(`turn\_function\_dial\_clockwise', 1) & Convection \\
(`turn\_function\_dial\_clockwise', 1) & Rotary \\
(`turn\_function\_dial\_clockwise', 1) & Off \\
(`turn\_function\_dial\_clockwise', 1) & Fermentation \\
(`turn\_function\_dial\_clockwise', 1) & Lower heater \\
\bottomrule
\end{tabularx}

\vspace{0.5em}
\textbf{Inferred Variable Definition:}

\begin{itemize}
    \item \textbf{Name:} \texttt{variable\_function}
    \item \textbf{Type:} \texttt{DiscreteVariable}
    \item \textbf{Value Range:}
    \begin{minted}[fontsize=\small, breaklines, breakanywhere]{python}
['Fermentation', 'Lower heater', 'Upper heater',
 'Lower & upper heater', 'Convection', 'Rotary', 'Off']
    \end{minted}
    \item \textbf{Current Value:} \texttt{'Off'}
\end{itemize}

\vspace{0.5em}
\textbf{Generated Code:}

\begin{minted}[fontsize=\small, breaklines, breakanywhere]{python}
variable_function_knob = DiscreteVariable(
    value_range=[
        'Fermentation', 'Lower heater', 'Upper heater',
        'Lower & upper heater', 'Convection', 'Rotary', 'Off'
    ],
    current_value='Off'
)
\end{minted}
\end{tcolorbox}

\newpage
\section{Details of Real World System}
\label{ap:realworld}

\subsection{Real World System Design}

 \begin{figure}[h]
     \centering
     \includegraphics[width=1.0\linewidth]{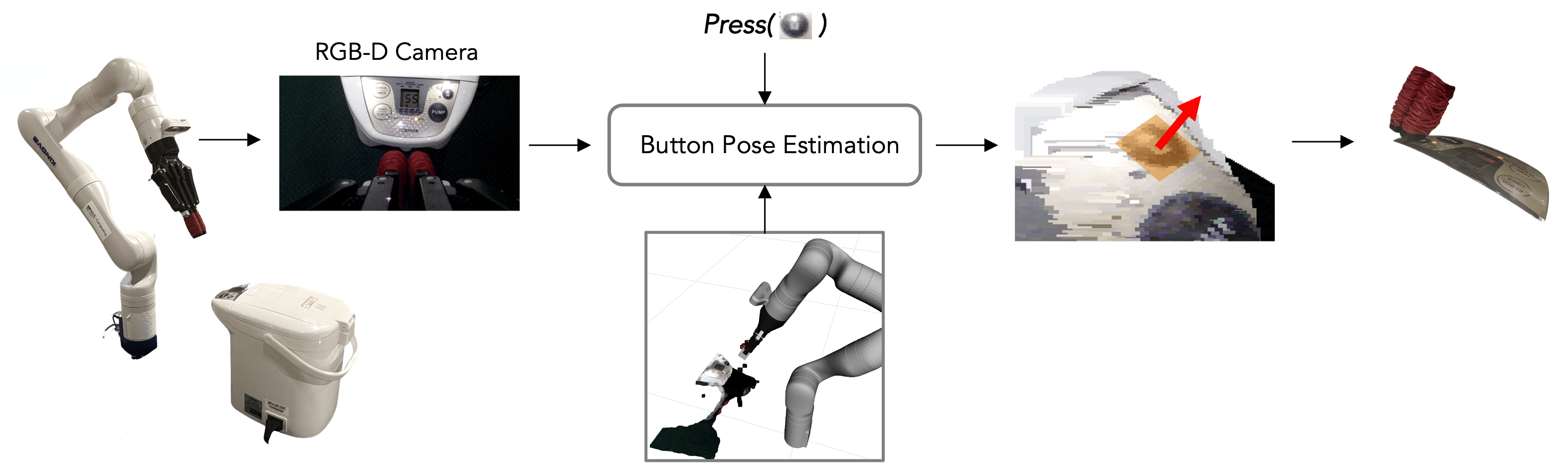}
     \vspace{-10pt}
     \caption{The real-world framework}
     \label{appendix:real-world}
     \vspace{-10pt}
 \end{figure}

In our real-world robotic system, we implement a framework that enables a manipulator to interact with physical appliances by pressing buttons accurately and robustly, as illustrated in Figure~\ref{appendix:real-world}. An RGB-D camera mounted near the robot’s end-effector captures both RGB images and depth data of the appliance interface. Given a press action parameterized by the bounding box of the target button, the button pose estimation module extracts the corresponding point cloud and computes the surface normal of the button region. This normal vector determines the correct approach angle for the robot to align its end-effector.

 \begin{figure}[h]
     \centering
     \includegraphics[width=0.9\linewidth]{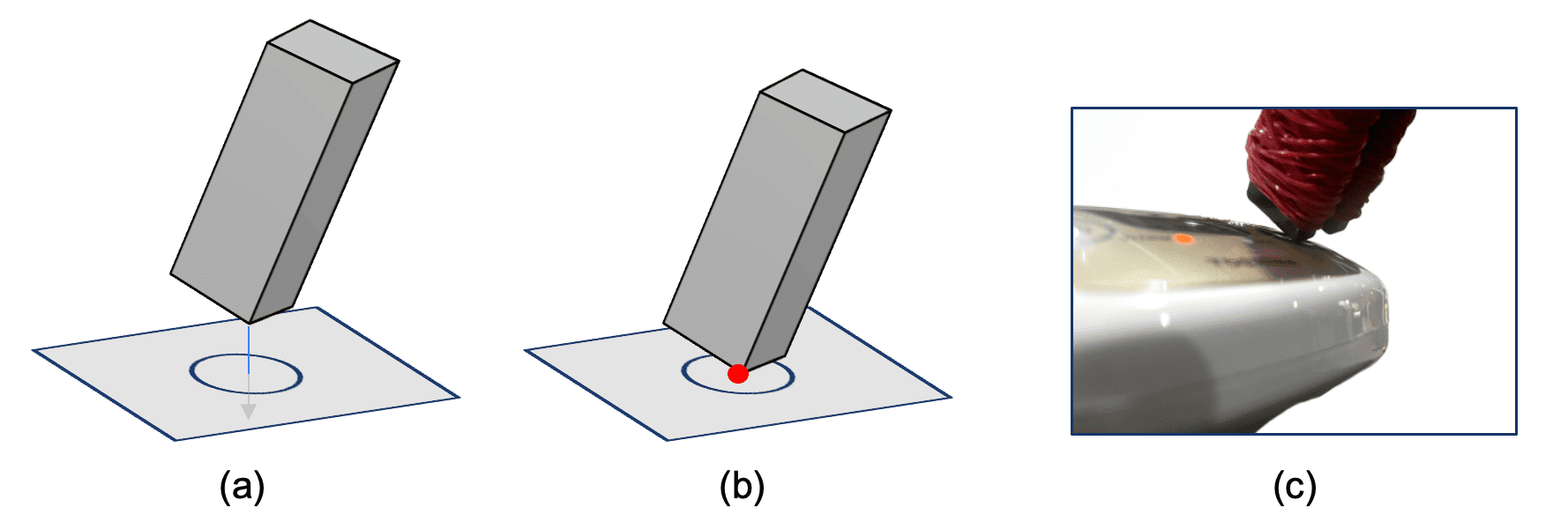}
     \vspace{-10pt}
     \caption{The pressing details}
     \label{appendix:press}
 \end{figure}

To reduce the contact area and improve precision, the robot aligns its gripper with the surface normal at a slight tilt. The pressing trajectory is generated in two stages: first, the end-effector moves to a position directly above the button; then, it advances 0.1 cm beyond the estimated button surface to ensure a firm press, as shown in Figure~\ref{appendix:press}. This approach compensates for minor depth inaccuracies and mechanical backlash, enhancing contact reliability.
The generated trajectory is executed using workspace tracking control, allowing the end-effector to follow the desired pressing motion precisely. This framework generalizes well across various devices and button types, demonstrating robustness to differences in button size, orientation, and mechanical resistance.

\subsection{Real World Experiments Setting}

 \begin{figure}[h]
     \centering
     \includegraphics[width=1\linewidth]{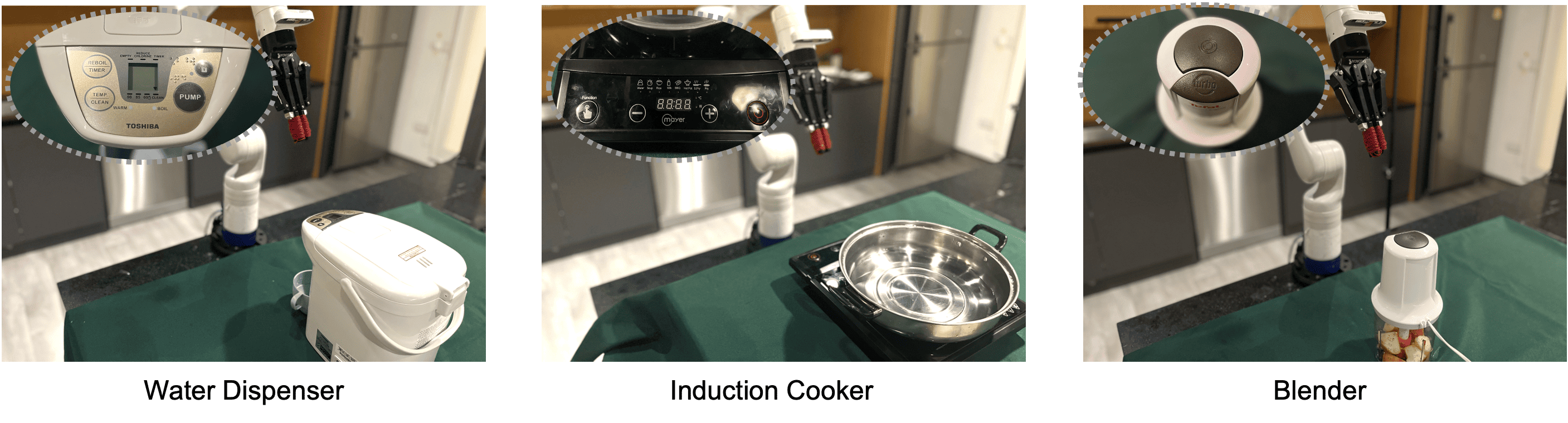}
     \vspace{-10pt}
     \caption{The real world setting}
     \label{appendix:realworldsetting}
 \end{figure}

To evaluate the performance and generalization ability of our proposed system, we conducted a series of real-world experiments involving common household appliances. Specifically, the robot was tasked with operating three distinct devices: a blender, an induction cooker, and a water dispenser, as shown in Fig. \ref{appendix:realworldsetting}. These appliances were selected for their diversity in interface design and physical interaction requirements, representing different types of button layouts, activation mechanisms, and task objectives.
For each appliance, we designed three task scenarios, resulting in a total of nine distinct interaction tasks. These tasks involve activating power buttons, selecting modes (e.g., milk mode or hot pot mode), or dispensing liquids, depending on the appliance. These tasks require the system to generalize based on visual input and prior knowledge for reasoning encoded in the user manual.

\begin{enumerate}[noitemsep, label=\textbf{T\arabic*.}, align=left, labelwidth=0.5cm]
    \item \textit{Select the HotPot mode and set power to 2000 W.}
    \item \textit{Select the Milk mode.}
    \item \textit{Select the HotPot mode and set power to 1600 W.}
    \item \textit{Set the insulation temperature to 98°, then pour the water.}
    \item \textit{Set the insulation temperature to 85°, then pour the water.}
    \item \textit{Set the insulation temperature to 65°, then pour the water.}
    \item \textit{Hold at slow speed for 10 seconds.}
    \item \textit{Hold at slow speed for 15 seconds.}
    \item \textit{Hold at turbo speed for 10 seconds.}
\end{enumerate}

\subsection{More Real World Execution Visualization}

\begin{figure*}[h]
  \centering
  \begin{tabular}{@{}c@{}}

  \begin{minipage}{\textwidth}
    \begin{flushleft}
    \textit{Select the HotPot mode and set power to 2000 W.}
    \end{flushleft}
    \includegraphics[width=0.8\textwidth]{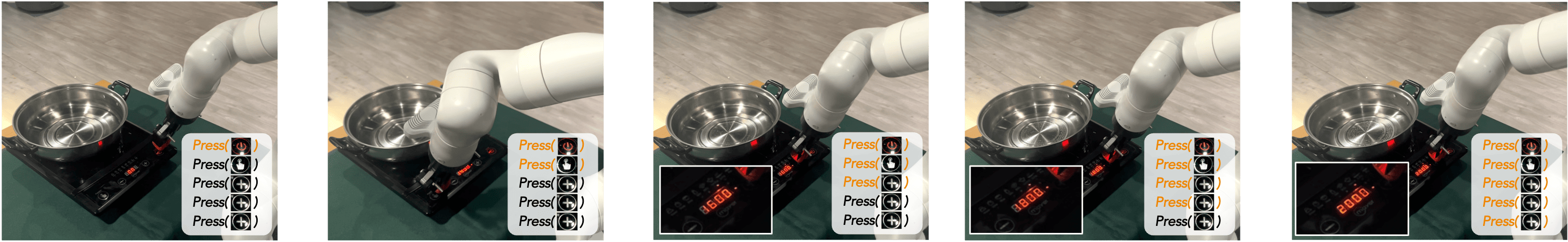}
  \end{minipage}
  \vspace{0.5em}
  \\

  \begin{minipage}{\textwidth}
    \begin{flushleft}
    \textit{Select the Milk mode.}
    \end{flushleft}
    \includegraphics[width=0.8\textwidth]{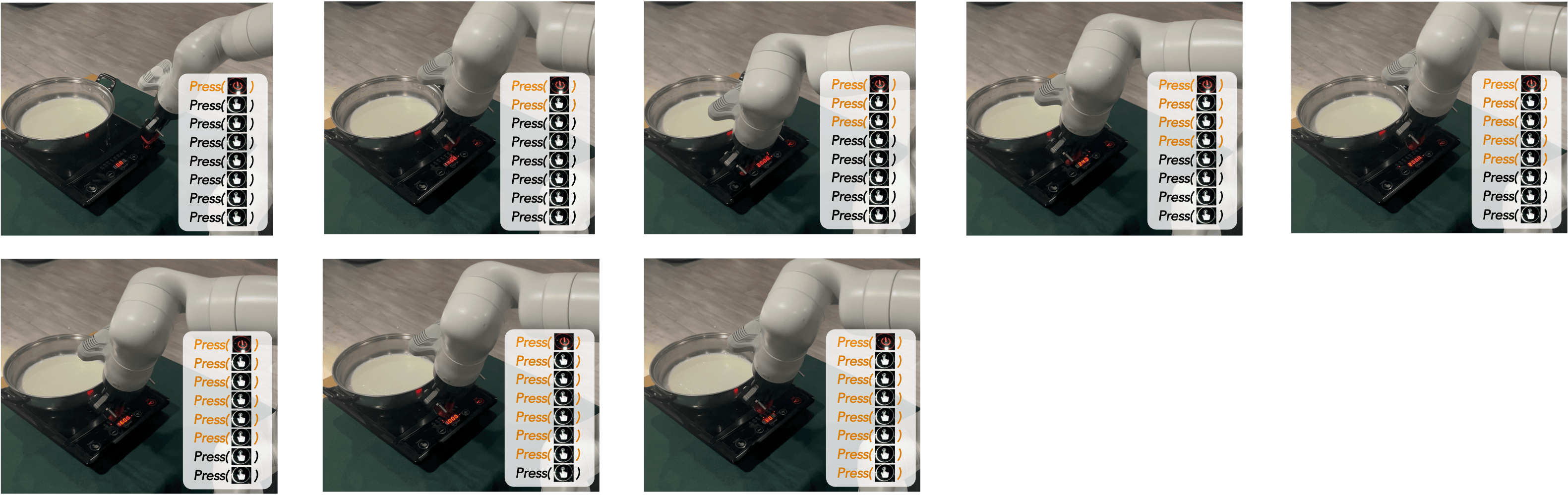}
  \end{minipage}
  \vspace{0.5em}
  \\

  \begin{minipage}{\textwidth}
    \begin{flushleft}
    \textit{Select the HotPot mode and set power to 1600 W.}
    \end{flushleft}
    \includegraphics[width=0.8\textwidth]{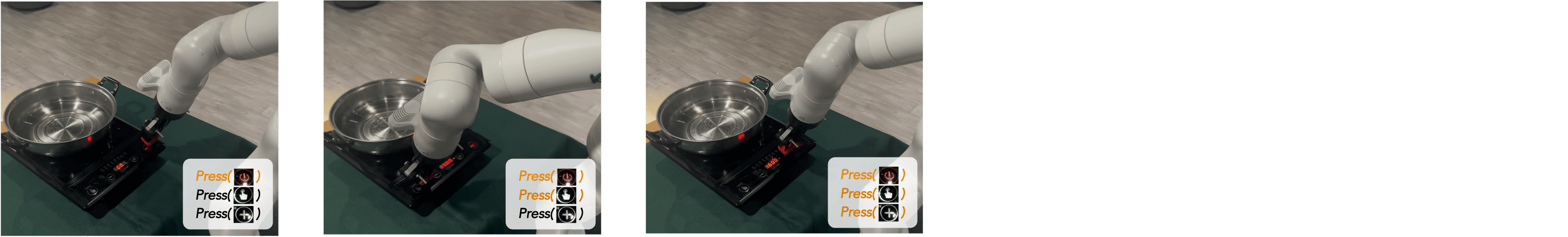}
  \end{minipage}
  \vspace{0.5em}
  \\

  \begin{minipage}{\textwidth}
    \begin{flushleft}
    \textit{Set the insulation temperature to 98°, then pour the water.}
    \end{flushleft}
    \includegraphics[width=0.8\textwidth]{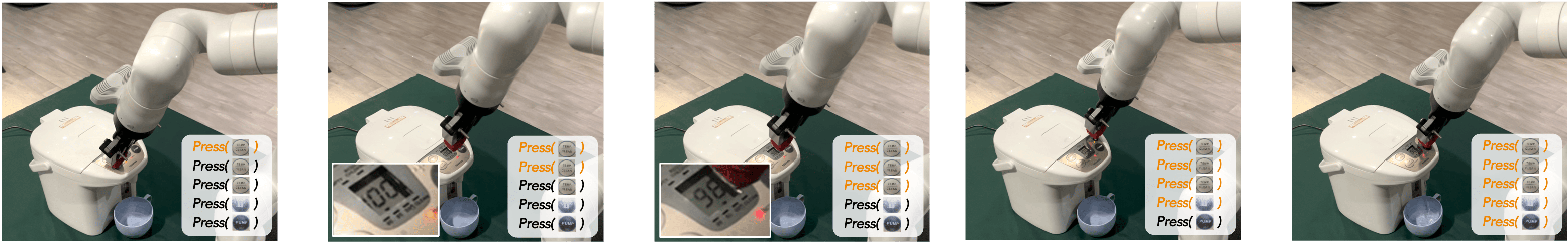}
  \end{minipage}
  \vspace{0.5em}
  \\

  \begin{minipage}{\textwidth}
    \begin{flushleft}
    \textit{Set the insulation temperature to 85°, then pour the water.}
    \end{flushleft}
    \includegraphics[width=0.8\textwidth]{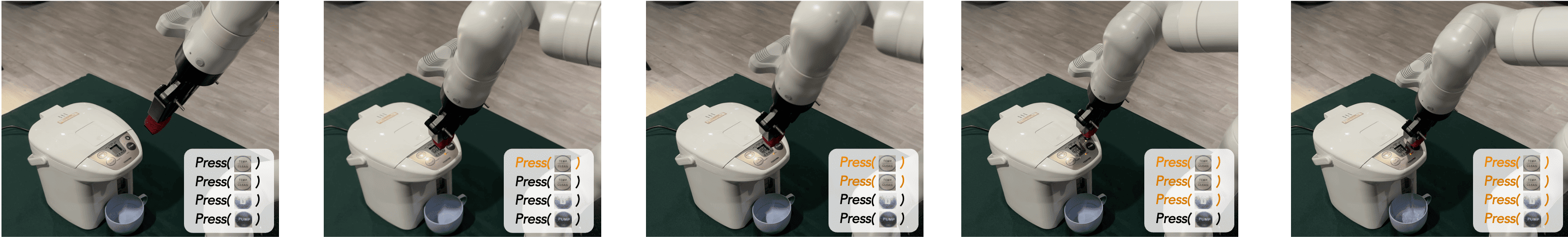}
  \end{minipage}
  \vspace{0.5em}
  \\

  \begin{minipage}{\textwidth}
    \begin{flushleft}
    \textit{Set the insulation temperature to 65°, then pour the water.}
    \end{flushleft}
    \includegraphics[width=0.8\textwidth]{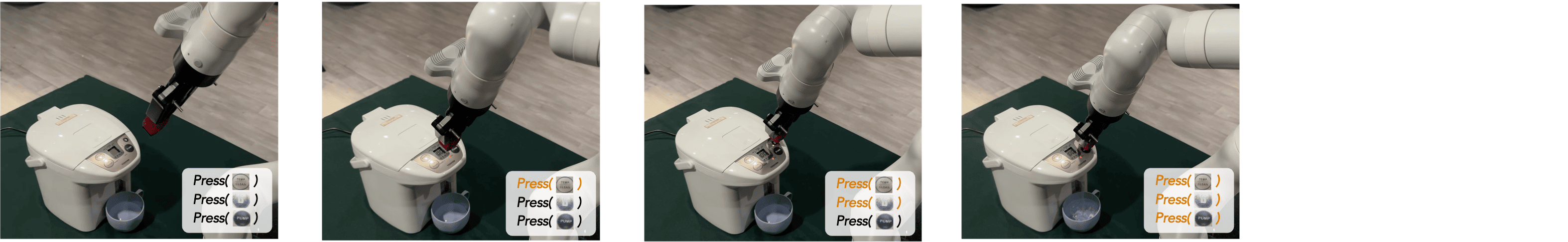}
  \end{minipage}
  \vspace{0.5em}
  \\

  \begin{minipage}{\textwidth}
    \begin{flushleft}
    \textit{Hold at slow speed for 10 seconds.}
    \end{flushleft}
    \includegraphics[width=0.8\textwidth]{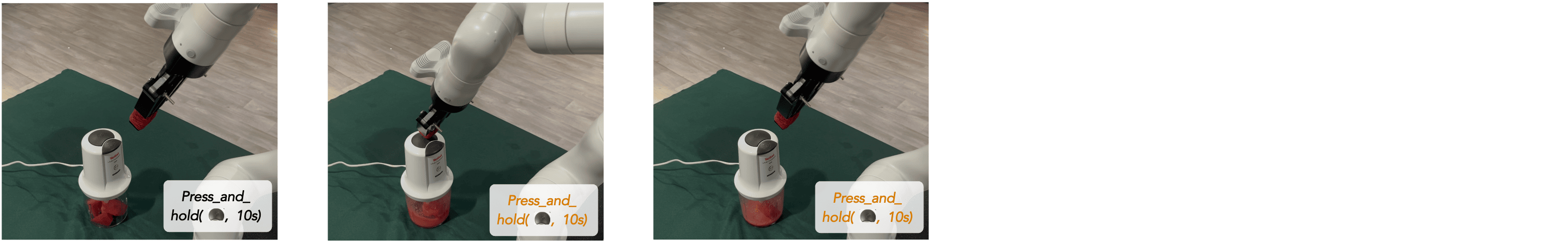}
  \end{minipage}
  \vspace{0.5em}
  \\

  \begin{minipage}{\textwidth}
    \begin{flushleft}
    \textit{Hold at slow speed for 15 seconds.}
    \end{flushleft}
    \includegraphics[width=0.8\textwidth]{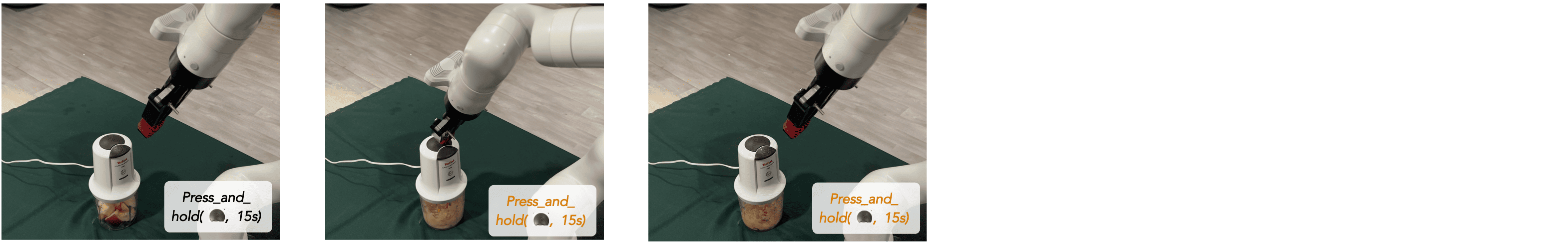}
  \end{minipage}
  \vspace{0.5em}
  \\

  \begin{minipage}{\textwidth}
    \begin{flushleft}
    \textit{Hold at turbo speed for 10 seconds.}
    \end{flushleft}
    \includegraphics[width=0.8\textwidth]{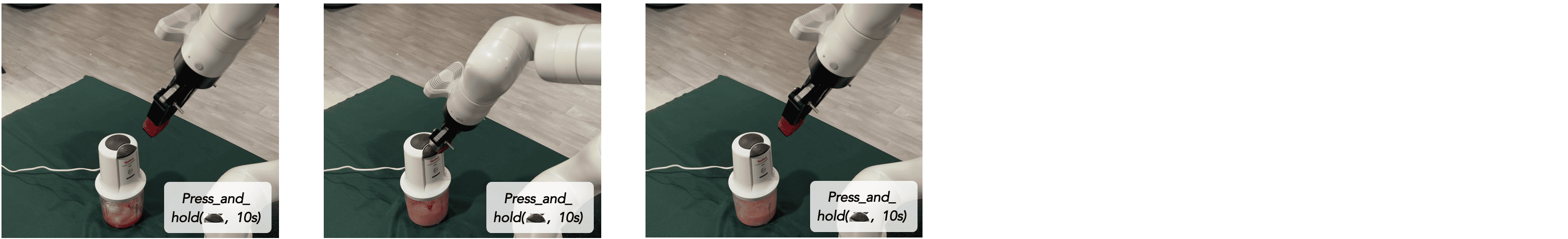}
  \end{minipage}

  \end{tabular}

  \caption{Snapshots of our system performing various tasks on real appliances. Each row shows execution steps for one task.}
  \vspace{-0.3cm}
  \label{fig:real_appliance_results_exs}
\end{figure*}

To evaluate the impact of parsing visual feedback from appliance displays, we built simulators for these three appliances, each using images of digital control panels to reflect state changes. For each appliance, 10 instructions were tested. LVLM agents were used to parsed the display after each action to infer the updated state, which was passed back as feedback to guide the next step. For \cotimage, the image was directly given to LVLM agents. For \myalg, feedback parsing is done by guiding the LVLM to focus on the variable currently being adjusted (Prompt~\ref{prompt:visual_feedback_parsing}). We compare \cotimage and \myalg in terms of success rate, execution steps, and SPL, as shown in Table~\ref{tab:real_world_stats}. Due to the simplicity of these appliances, both methods show similar performance. For more visualization, please see Fig. \ref{fig:real_appliance_results_exs}.

\begin{table}[t]
\centering
\caption{Detailed Performance of Success Rate / Execution Step / SPL by Appliance Types}
\vspace{0.5em}
\resizebox{\textwidth}{!}{%
\begin{tabular}{lcccc}
\toprule
\textbf{Method} & \textbf{Blender} & \textbf{Water Dispenser} & \textbf{Induction Cooker} & \textbf{Avg} \\
\midrule
\cotimage & 1.0 / 3.50 / 0.29 & 1.0 / 5.70 / 0.59 & 0.5 / 17.70 / 0.13 & 0.83 / 8.97 / 0.34 \\
\myalg    & 1.0 / 1.0 / 1.0   & 0.7 / 8.4 / 0.31   & 1.0 / 16.5 / 0.38  & 0.9 / 8.63 / 0.56 \\
\bottomrule
\end{tabular}%
}
\label{tab:real_world_stats}
\end{table}

\section{Details of Failure Mode Analysis}
\label{ap:failures}

We categorize and analyze the main failure modes observed across different baseline methods as shown below.

\paragraph{Failure due to Lack of Action Grounding}
This is defined as failures that occur due to incorrect grounding of actions to visual elements. For example, the model may select a neighboring button instead of the correct one because LVLMs struggle to associate OCR text labels with the correct control panel element. This highlights limitations in visual-text alignment within vision-language models. This failure mode is mainly applicable to: \cotimage.

\paragraph{Failure due to Lack of Structured Model} 

This mainly include failures that occur due to (1) incorrect association between actions and effects due to lack of transition modeling; (2) repeated adjustment of the same variable due to lack of macro actions; (3) premature ending or wrongly parsed visual feedback due to lack of state estimation; (4)  incorrect goal state specification due to lack of structured goal states. This failure mode is mainly applicable to \cotimage; \cotaction; \mynomodel.

\paragraph{Failure due to Incorrect Transition Model}

This mainly includes failures caused by incorrect interpretation of transition rules, especially when the variable exhibits irregular step sizes in its value space. This failure mode is mainly applicable to: \mynoreason; \mynoupdate.

\paragraph{Failure due to Hallucinated Model Details} 
This includes failures caused by LLM hallucination during model construction, resulting in invalid transition rules that fail syntax checks. This failure mode is mainly applicable to: \myalg, \mynoreason, \mynomodel, \mynoupdate.

\section{Prompts}
\label{ap:prompts}
In this section, we provide the detailed prompts for two baselines: \cotimage (Prompt \ref{prompt:cotimage}) and \cotaction (Prompt \ref{prompt:cotaction}).
The remaining ablation methods share the same prompts as \myalg.

For \myalg, we provide prompts for three sections: (1) Build appliance models; (2) Update appliance models using closed-loop feedback; (3) Action grounding.
To build appliance models, we need to (1) extract control panel element names (Prompt~\ref{prompt:extract_control_panel_element_names}) and action names (Prompt~\ref{prompt:extract_action_names}); (2) extract variables (Prompt~\ref{prompt:extract_variables}), macro actions (Prompt~\ref{prompt:extract_features}), and generate the appliance model with extracted information (Prompt~\ref{prompt:extract_appliance_model}); finally (3) 
Generate task policy and goal state based on the appliance model (Prompt ~\ref{prompt:generate_task_policy_and_goal_state}). 

To update the appliance model using closed-loop feedback, the steps include: (1) 
After execution of each macro action and receiving feedback, \myalg parses the feedback (Prompt~\ref{prompt:visual_feedback_parsing}) and compares the goal with the feedback (Prompt~\ref{prompt:compare_goal_state_with_feedback}).
If the goal is achieved, it proceeds to the next action. Otherwise, it executes exploration actions to collect a sequence of observations. Then, it uses them to diagnose the incorrectly modeled variable (Prompt~\ref{prompt:diagnose_incorrect_variable_definition}),
updates the variable definition (Prompt~\ref{prompt:update_variable_definition}),
and updates the appliance model (Prompt~\ref{prompt:update_appliance_model}) and goal state (Prompt~\ref{prompt:update_goal}) accordingly.  

To perform action grounding, we need to: 
(1) Use LVLMs to detect candidate bounding boxes for control panel elements, then remove false positives using LVLMs (Prompt~\ref{prompt:check_bbox_contain_control_panel_elements}). This step ensures only valid regions are kept before passing them for slower, more detailed grounding.
(2) Map bounding boxes to control panel element names (Prompt~\ref{prompt:map_bbox_to_names}).
(3) Remove duplicate bounding boxes being mapped to the same control panel element (Prompt~\ref{prompt:remove_duplicate_bbox}).
And (4) Map each action name to a grounded control panel element name and an action type (Prompt~\ref{prompt:ground_actions}).

\begin{promptbox}[label=prompt:cotimage]{\cotimage Action Proposal}
\small

You are given:
\begin{itemize}
    \item Two images: 
    \begin{enumerate}[label=(\arabic*)]
        \item A photo of the appliance control panel.
        \item A version with indexed bounding boxes circling the control panel elements (buttons, dials).
    \end{enumerate}
    \item A user command describing the target task.
    \item User manual.
    \item A set of allowed action types: \texttt{press}, \texttt{hold}, \texttt{turn\_dial\_clockwise}, \texttt{turn\_dial\_anti\_clockwise}.
    \item Optionally, display panel feedback in text after each action.
\end{itemize}

\vspace{0.5em}
\textbf{Action Proposal Rules:}
\begin{itemize}

    \item At the start of the task, assume the initial appliance state is unknown. Execute an action to receive feedback. On subsequent steps, use observed display panel feedback to reason about the current state, and propose the next action needed to complete the task.
    \item Only one action is allowed per response, but you can execute it multiple times (e.g., set \texttt{execution\_times = 2}.
    \item \texttt{hold} actions require specifying a duration. If not mentioned in manual, default to 10 seconds. \texttt{hold} can involve two buttons simultaneously and requires a \texttt{duration}. The other action types apply to a single button or dial.
    \item If the task is completed or infeasible (e.g., display feedback remains wrong after repeated attempts failed), return an \texttt{end} action to stop. 
\end{itemize}

\vspace{0.5em}
\textbf{Output Format:} Return 5 Python variables in the following format:
\begin{tcolorbox}[colback=gray!5]
\begin{minted}[breaklines, breakanywhere]{python}
variable_reason = "<Your reasoning>"
action_type = "press_button"  # or other valid type
bbox_index = 5                 # int or [int, int] if pressing two buttons
execution_times = 1           # integer count
duration = None               # duration in seconds if hold, otherwise None

# to terminate a task:
variable_reason = "Task is completed / unable to achieve."
action_type = "end" 
bbox_index = None
execution_times = None 
duration = None
\end{minted}
\end{tcolorbox}

\vspace{0.5em}
\textbf{Example:} 
\begin{tcolorbox}[colback=gray!5]
\begin{minted}[breaklines, breakanywhere]{python}
# User instruction: Set the dial (index = 8) from \texttt{OFF} to \texttt{3}.
variable_reason = "Current power value is OFF. I will turn the dial clockwise 3 times to set it to 3."
action_type = "turn_dial_clockwise"
bbox_index = 8
execution_times = 3
duration = None
\end{minted}
\end{tcolorbox}

\end{promptbox}

\begin{promptbox}[label=prompt:cotaction]{\cotaction Action Proposal}
\small

You are given:
\begin{itemize}
    \item A user command describing the target task.
    \item User manual.
    \item A list of available executable actions.
    \item Optionally, display panel feedback in text after each action.
\end{itemize}

\vspace{0.5em}
\textbf{Action Proposal Rules:}
\begin{itemize}
    \item At the start of the task, assume initial appliance state is unknown. Execute an action to receive feedback. On subsequent steps, use observed display panel feedback to reason about current state, and propose the next action needed to complete the task.
    \item Use only the listed available actions. Each action should be returned as a Python function call. Provide a clear and concise reason using \texttt{variable\_reason}.
    \item Only one action is allowed per response, but you can execute it multiple times (e.g., set \texttt{execution\_times = 3}).
    \item If a \texttt{hold} action causes values to change too quickly, avoid using it. Use repeated \texttt{press} actions instead. \texttt{hold} actions require specifying a duration. If not mentioned in the manual, default to 10 seconds. 
    \item If the task is completed or infeasible (e.g., display feedback remains incorrect after repeated attempts), return an \texttt{end} action to stop.
\end{itemize}

\vspace{0.5em}
\textbf{Output Format:} Return 2 Python variables in the following format:
\begin{tcolorbox}[colback=gray!5]
\begin{minted}[breaklines, breakanywhere]{python}
variable_reason = "<Your reasoning>"
variable_response_string = "run_action('action_name', execution_times=N)"

# Example of hold actions:
variable_response_string = "run_action('hold_buttonX_and_buttonY', execution_times=1, duration=5)" # 5 seconds

# To terminate the task:
variable_reason = "Task is completed / unable to achieve."
variable_response_string = "end"
\end{minted}
\end{tcolorbox}

\vspace{0.5em}
\textbf{Example:} 
\begin{tcolorbox}[colback=gray!5]
\begin{minted}[breaklines]{python}
# User instruction: Set the dial from OFF to 3 by turning it clockwise.
variable_reason = "Current power value is OFF. I will turn the dial clockwise 3 times to set it to 3."
variable_response_string = "run_action('turn_dial_clockwise', execution_times=3)"
\end{minted}
\end{tcolorbox}
\end{promptbox}

\begin{promptbox}[label=prompt:extract_control_panel_element_names]{Extract Control Panel Element Names}
\small
You are given an appliance user manual and an image of its control panel. Identify all \textbf{control panel elements}, i.e., \texttt{button} and \texttt{dial}.

\vspace{0.5em}
\textbf{Identification Guidelines:}
\begin{itemize}
    \item Include elements mentioned in the manual or shown in the image if they clearly correspond to a described function.
    \item Use one name per physical control. If it adjusts multiple settings, use a combined name\\
    (e.g., \texttt{power\_timer\_dial}, not \texttt{power\_dial} and \texttt{timer\_button}). If the manual names a button (e.g., \texttt{function\_button}), use that name, even if the image shows only labels of its configurations like \texttt{menu 1}, \texttt{menu 2}, \texttt{menu 3}.
    \item List each distinct button separately, even if they adjust the same function. Examples: \texttt{air\_roast\_button}, \texttt{air\_fry\_button}; \texttt{increase\_button}, \texttt{decrease\_button}; \texttt{number\_0\_button}, \texttt{number\_1\_button}, ...

\end{itemize}

\vspace{0.5em}
\textbf{Exclude:}
\begin{itemize}
    \item Non-executable parts such as printed labels, static icons, light indicators, and digital displays.
    \item Any component not on the control panel, such as power plugs or lids.
\end{itemize}

\vspace{0.5em}
\textbf{Naming Conventions:}
\begin{itemize}
    \item Use \texttt{name\_type} format (e.g., \texttt{start\_stop\_button}, \texttt{power\_level\_dial}). 
    \item Only lowercase letters, digits, and underscores are allowed. No spaces or special characters.
\end{itemize}

\vspace{0.5em}
\textbf{Output Format:}
\begin{itemize}
    \item Return a Python list named \texttt{names\_list}.
    \item Each item must be a string with a Python comment describing its function, location, and any visible symbol (e.g., triangle, bottle, arrow).
\end{itemize}

\vspace{0.5em}
\textbf{Example Output:}
\begin{tcolorbox}[breakable, colback=gray!5]
\begin{minted}[breaklines, breakanywhere]{python}
names_list = [
    "start_stop_button",  # starts/stops cooking; lower right; triangle icon
    "number_1_button",    # sets time; middle keypad; labeled '1'
    "increase_button",    # increases value; top left; '+' symbol
]
\end{minted}
\end{tcolorbox}

\end{promptbox}

\begin{promptbox}[label=prompt:extract_action_names]{Extract Action Names}
\small
You are given an appliance user manual and a list of \textbf{control panel element} names. Your task is to identify all \textbf{executable actions} that are:

\begin{enumerate}[label=(\arabic*)]
    \item \textbf{described in the user manual}, and
    \item Involve \textbf{control panel elements} listed above (e.g., buttons, dials).
\end{enumerate}

\vspace{0.5em}
Carefully match each control element with relevant actions described in the manual.

\vspace{0.5em}
\textbf{Valid Action Types:}
\begin{itemize}
    \item \texttt{press\_<element\_name>}
    \item \texttt{hold\_<element\_name>} \hfill \#(duration = x seconds; use 3 if unspecified)
    \item \texttt{hold\_<element1>\_and\_<element2>} \hfill  \#(duration = x seconds; use 3 if unspecified)
    \item \texttt{turn\_<element\_name>\_clockwise} \hfill (only valid for dials)
    \item \texttt{turn\_<element\_name>\_anticlockwise} \hfill (only valid for dials)
\end{itemize}

\vspace{0.5em}
\textbf{Naming conventions:}
\begin{itemize}
    \item Construct each action by selecting a valid action type from the list above and inserting a control element name from the provided list. 
    \item Use lowercase letters, digits, and underscores only. Do not include any special characters or symbols.
    
\end{itemize}

\vspace{0.5em}
\textbf{Exclusions:}
\begin{itemize}
    \item Do not include actions not mentioned in the manual.
    \item Do not create duplicate or ambiguous actions.
    \item Do not include duration in the action name. Write it as a comment on the same line.
\end{itemize}

\vspace{0.5em}
\textbf{Output Format:}
List each valid action as a separate line of plain text.

\vspace{0.5em}
\textbf{Example Output:}
\begin{tcolorbox}
\begin{minted}{text}
press_kitchen_timer_button
press_time_dial
press_and_hold_stop_button  #(duration = 5 seconds)
press_and_hold_start_button_and_cancel_button  #(duration = 3 seconds)
turn_power_level_dial_clockwise
turn_power_level_dial_anticlockwise
\end{minted}
\end{tcolorbox}
\end{promptbox}

\begin{promptbox}[label=prompt:extract_variables]{Extract Variables}
\small
You are given an appliance user manual, a list of executable action names, a list of control panel element names, and a list of predefined variable classes in Python. Your task is to extract all appliance \textbf{variables} as instances of the predefined Python classes.

\vspace{0.5em}
\textbf{Definition of Variable:}
An internal configuration state of the appliance that can be adjusted through actions (e.g., power level, temperature, time).

\vspace{0.5em}
\textbf{How to Identify a Variable:}
User manuals often describe multiple \textbf{features} (i.e. high-level functions like \texttt{Defrost}, \texttt{Grill}), each consisting of actions that configure internal appliance states. These states are the \textbf{variables}. For example, a microwave may include \texttt{Defrost} and \texttt{Grill} features, both of which adjust \texttt{menu} and \texttt{time}, but assign different values depending on the feature. Here, \texttt{Defrost} and \texttt{Grill} are features. \texttt{menu} and \texttt{time} are variables shared across features. Define a variable if:

\begin{enumerate}[label=(\arabic*)]
\item It is explicitly \textbf{described in the manual},
\item It is adjusted via a \textbf{listed control panel element name} (e.g., button, dial), and
\item It is modified by an \textbf{listed action action}.
\end{enumerate}

\vspace{0.5em}
\textbf{Naming Convention:} Use the format \texttt{variable\_<variable\_name>}. Use only lowercase letters and underscores.

\begin{tcolorbox}[breakable, colback=gray!5]
\begin{minted}[breaklines, breakanywhere]{python}
variable_power_on_off = ... # User manual: Press POWER to turn off.
variable_child_lock = ...
variable_start_pause = ...
\end{minted}
\end{tcolorbox}

\vspace{0.5em}
\textbf{Valid Variable Types:} Used to define variable transition rules. Each variable type can be directly invoked via code. Each variable can have its value changed by \texttt{.next()} and \texttt{.prev()} or directly assigned by \texttt{.set\_current\_value()}. 
\vspace{0.5em}
\begin{enumerate}[label=(\arabic*)]
\item \texttt{DiscreteVariable}: Categorical values. Value range consists of strings.
\begin{tcolorbox}[breakable, colback=gray!5]
\begin{minted}[breaklines, breakanywhere]{python}
variable_power = DiscreteVariable(value_range=["on", "off"], current_value="on")
variable_mode = DiscreteVariable(value_range=["eco", "turbo", "auto"], current_value="eco")
\end{minted}
\end{tcolorbox}

\item \texttt{ContinuousVariable}: Numerical values. Supports piecewise ranges.
\begin{tcolorbox}[breakable, colback=gray!5]
\begin{minted}[breaklines, breakanywhere]{python}
variable_clock_setting_hour = ContinuousVariable(value_ranges_steps=[[0, 23, 1]], current_value=0) # value range: 0-23 hours, step size: 1 hour
variable_wash_time = ContinuousVariable(value_ranges_steps=[[0, 3, 3], [3, 15, 1]], current_value=0) # value range: 0 or 3-15 minutes
\end{minted}
\end{tcolorbox}

\item \texttt{TimeVariable}. Supports "hour-minute-second" format. 
\begin{tcolorbox}[breakable, colback=gray!5]
\begin{minted}[breaklines, breakanywhere]{python}
variable_timer = TimeVariable(values_ranges_steps = [('00:00:00', '00:59:00', 60)], current_value='00:00:00') # value range: 0-59 minutes: step size: 1 min
\end{minted}
\end{tcolorbox}

\item \texttt{InputString}. Stores keypad input sequence. 
\begin{tcolorbox}[breakable, colback=gray!5]
\begin{minted}[breaklines, breakanywhere]{python}
# User manual: Enter a 3-digit code using number pads to set the timer. 
variable_input_string = InputString()
\end{minted}
\end{tcolorbox}
\end{enumerate}
\vspace{0.5em}
\textbf{Output Format:}
Executable python code that defines each variable. The current variable value should be initialised to the first value in the range if not otherwise specified by the manual. 

\vspace{0.5em}

\textbf{Example Output:}
\begin{tcolorbox}
\begin{minted}[breaklines, breakanywhere]{python}
variable_power = DiscreteVariable(value_range = ["on", "off], current_value = "on")
variable_temperature = ContinuousVariable(value_ranges_stpes = [[20, 30, 1]], current_value = 20)
\end{minted}
\end{tcolorbox} 

\textbf{Special Cases:}

\begin{enumerate}[label=(\arabic*)]
\item \textbf{Setting Adjustable via Different Features:}
If a setting can be adjusted in different features using different transition rules (i.e. how a variable's value changes given an action), define a separate variable for each (e.g., \texttt{cook\_time} set via number pads vs. incremented by \texttt{press\_start\_button}). 

\begin{tcolorbox}[breakable, colback=gray!5]
\begin{minted}[breaklines, breakanywhere]{python}
# User manual <normal cook>:
# 1) Press "COOK" once;
# ...
# 4) Use the number pads to enter cooking time in MM:SS format (e.g., to set 6 minutes, press "6", "0", "0");
# 5) Press "COOK" again to confirm.
variable_normal_cook_time = ... 

# User manual <speedy cook>:
# 1) Press "Start" to start cooking for 30 seconds. Each subsequent press adds time by 30 seconds.  
variable_speedy_cook_time = ... 
\end{minted}
\end{tcolorbox}

\item \textbf{Setting Adjusted Across Different Feature Steps:}
If a setting is adjusted in multiple steps (e.g., hour and minute of a timer) in a feature, define one variable per step.
\begin{tcolorbox}[breakable, colback=gray!5]
\begin{minted}[breaklines, breakanywhere]{python}
# User manual <clock setting>:
# 1) Press "CLOCK" once, the hour figure flashes.
# 2) Press "up arrow" or "down arrow" to adjust the hour (0--23).
# 3) Press "CLOCK", the minute figure flashes.
# 4) Press "up arrow" or "down arrow" to adjust the minute (0--59).
# 5) Press " CLOCK " to finish clock setting. ":" will flash, the "clock symbol" indicator will go out. The clock setting has been finished.
variable_clock_setting_hour = ... 
variable_clock_setting_minute = ... 
\end{minted}
\end{tcolorbox}

\item \textbf{Setting Conditioned on Program Choice:} If a setting's value range depends on the selected program, (e.g. microwave menu, washing machine program), follow this structure. 
{\footnotesize

\begin{itemize}
    \item Define a selector variable, e.g., \texttt{variable\_program\_index}, to store the chosen program.
    \item Define a placeholder variable, e.g., \texttt{variable\_program\_setting = None}, which is dynamically assigned.
    \item For each program, define a separate variable using the format \texttt{variable\_<feature\_name>\_<program\_name>} (e.g., \texttt{variable\_set\_program\_popcorn}).
    \item Create a dictionary \texttt{program\_setting\_dict} to map each program to its respective setting variable.
\end{itemize}
}
\begin{tcolorbox}[breakable, colback=gray!5]
\begin{minted}[breaklines, breakanywhere]{python}
# User manual:
# Microwave program popcorn sets size (1 cup, 2 cup), pizza sets weight (250g, 350g, 450g), soup sets volume (200ml, 300ml, 400ml).
# Each time a new program is selected, variable_program_setting is updated using program_setting_dict.

# variable A (selector)
variable_program_index = DiscreteVariable(["popcorn", "pizza", "soup"], "popcorn")

# variable B (placeholder)
variable_program_setting = None

# program-specific variables
variable_program_setting_popcorn = DiscreteVariable(["1 cup", "2 cup"], "1 cup")
variable_program_setting_pizza = DiscreteVariable(["250g", "350g", "450g"], "250g")
variable_program_setting_soup = DiscreteVariable(["200ml", "300ml", "400ml"], "200ml")

# mapping dictionary
program_setting_dict = {
    "popcorn": variable_program_setting_popcorn,
    "pizza": variable_program_setting_pizza,
    "soup": variable_program_setting_soup
}
# Selecting a mode updates variable_menu_setting from this dictionary.
\end{minted}
\end{tcolorbox}

\end{enumerate}

\end{promptbox}

\begin{promptbox}[label=prompt:extract_features]{Extract Features}
\small

You are given the user manual of an appliance, a list of executable action names, a list of variables, and a predefined \texttt{Feature()} class in Python. Your task is to extract all appliance \textbf{features} as an instance of the predefined \texttt{Feature()} object.

\vspace{0.5em}
\textbf{Definition of Feature:} A high-level operation (e.g., \texttt{clock setting}, \texttt{cooking}) consisting of step-by-step procedures that adjust one or more variables using valid actions.

\vspace{0.5em}
\textbf{Output Format:} Define a dictionary \texttt{feature\_list}, where each item is a feature name and its value is a list of steps. Each step is a dictionary with:
\begin{enumerate}[label=(\arabic*)]
  \item \texttt{step} index (integer),
  \item \texttt{actions} (list of action strings),
  \item Optional \texttt{variable} adjusted in this step,
  \item Optional \texttt{comment} describing fixed action effects or input string parsing requirements.
\end{enumerate}
\vspace{0.5em}
If any actions or variables are unused, include them under the reserved feature \texttt{"null"}:
\begin{tcolorbox}[colback=gray!5]
\begin{minted}[breaklines, breakanywhere]{python}
feature_list["null"] = [{"step": 1, "actions": ["unused_action_1"], "missing_variables": ["variable_a"]}]
\end{minted}
\end{tcolorbox}

\vspace{0.5em}
Conclude with:
\begin{tcolorbox}[colback=gray!5]
\begin{minted}[breaklines, breakanywhere]{python}
simulator_feature = Feature(feature_list=feature_list, current_value=("empty", 1))
\end{minted}
\end{tcolorbox}

\vspace{0.5em}
\textbf{Example Output:}
\begin{tcolorbox}[colback=gray!5]
\begin{minted}[breaklines, breakanywhere]{python}
# User manual <clock setting>:
# 1) Press "CLOCK" once, the hour figure flashes.
# 2) Press "up arrow" or "down arrow" to adjust the hour (0--23).
# 3) Press "CLOCK", the minute figure flashes.
# 4) Press "up arrow" or "down arrow" to adjust the minute (0--59).
# 5) Press " CLOCK " to finish clock setting. ":" will flash, the "clock symbol" indicator will go out. The clock setting has been finished.

feature_list = {}
feature_list["clock_setting"] = [
    {"step": 1, "actions": ["press_clock_button"]}, 
    {"step": 2, "actions": ["press_up_arrow_button", "press_down_arrow_button"], "variable": "variable_clock_setting_hour"},
    {"step": 3, "actions": ["press_clock_button"]}, 
    {"step": 4, "actions": ["press_up_arrow_button", "press_down_arrow_button"], "variable": "variable_clock_setting_minute"},
    {"step": 5, "actions": ["press_clock_button"]} 
]
feature_list["null"] = [{"step": 1, "actions": [],
"missing_variables": []}]
simulator_feature = Feature(feature_list=feature_list, current_value=("empty", 1))
\end{minted}
\end{tcolorbox}

\vspace{0.5em}
\textbf{Identification Guidelines:}
\begin{enumerate}[label=(\arabic*)]
  \item Only model features with clear step-by-step instructions written in the user manual. Ignore features introduced only by naming buttons and dials without full procedures.
  \item Exclude non-essential features like WiFi, app control, remote control, reset, cleaning, multi-stage cooking, sound/audio settings, memory, touchscreen feedback, or progress queries after operation starts. For \texttt{hold\_<element>} actions, ignore action effects that merely speed up changes. Only model a \texttt{hold} action if it toggles a function (e.g., child lock).
 \item  Split features into shorter, reusable units where possible. For \textit{consecutive steps in a feature}, if they adjust different variables, consider separating them into distinct features (e.g. \texttt{start}, \texttt{cancel}, \texttt{power\_on}). If \textit{consecutive steps in a feature} adjust the same variable (e.g., \texttt{lock}/\texttt{unlock}), merge them.
 \item The feature that should stay merged is program settings, as the specific program setting is conditioned the program choice (e.g. \texttt{pizza} program requires setting \texttt{cooking\_weight}, but \texttt{soup} program requires setting \texttt{soup\_volume} (explained in \textit{extract variable}). Follow this structure: 
 \begin{tcolorbox}[colback=gray!5]
\begin{minted}[breaklines, breakanywhere]{python}
feature_list["set_program"] = [
  {"step": 1, "actions": ["press_program_button"], "variable": "variable_program_index"},
  {"step": 2, "actions": ["press_plus_button", "press_minus_button"], "variable": "variable_program_setting"}
]
  \end{minted}
\end{tcolorbox} 
 
  \item If an action always sets a variable to a fixed value, remark in \texttt{"comment"}. 
  \begin{tcolorbox}[colback=gray!5]
  \begin{minted}[breaklines, breakanywhere]{python}
feature_list["start_cooking"] = {"step": 1, "actions": ["press_start_button"], "variable": "variable_start_cooking", 
 "comment": "start always set to on"}
  \end{minted}
  \end{tcolorbox}
\item If an action affects multiple variables, set the variable whose values will be assigned dynamically under \texttt{variable} and describe those with fixed target values in \texttt{comment}. 
  \begin{tcolorbox}[colback=gray!5]
  \begin{minted}[breaklines, breakanywhere]{python}
  # user manual <speedy cooking>: 
  # press start button will immediately start cooking at 100% power for 30 seconds. Each subsequent press increases cooking time by 30 seconds. 
feature_list["start_cooking"] = {"step": 1, "actions": ["press_start_button"], "variable": "variable_cooking_time", 
 "comment": "variable_start set to on, variable_power set to 100"}
  \end{minted}
  \end{tcolorbox}
  \item \texttt{turn\_dial} actions must match both direction and effect. If \texttt{turn\_dial} affects different variables in different directions, distinguish them (e.g., clockwise for \texttt{time}, anticlockwise for \texttt{power}).
    \begin{tcolorbox}[colback=gray!5]
  \begin{minted}[breaklines, breakanywhere]{python}
feature_list["adjust_time"] = [{"step": 1, "actions": ["turn_dial_clockwise"], "variable": "variable_time"}]
feature_list["adjust_power"] = [{"step": 1, "actions": ["turn_dial_anticlockwise"], "variable": "variable_power"}]
  \end{minted}
    \end{tcolorbox}

  \item To compactly describe appliance features that input values via number pads, you can use the given \texttt{meata\_actions\_on\_numbers} to refer all the number pads, and track them with \texttt{meta\_actions\_dict}. Make a comment beside the variable whose value assignment requires parsing from input string.
  \begin{tcolorbox}[colback=gray!5]
\begin{minted}[breaklines, breakanywhere]{python}
# Predefined
meta_actions_on_number = [
    "press_number_0_button", "press_number_1_button", ..., "press_number_9_button"
]
meta_actions_dict = {
    "0": "press_number_0_button",
    "1": "press_number_1_button",
    ...
}

# Example usage
feature["set_timer"] = [
{"step": 1, "actions": ["press_timer_button"],},
{"step": 2, "actions": meta_actions_on_numbers, "variable": "variable_timer", 
 "comment": "requires parsing from variable_input_string"}]
\end{minted}
\end{tcolorbox}
\end{enumerate}
\end{promptbox}

\begin{promptbox}[label=prompt:extract_appliance_model]{Extract Appliance Model}
\small

You are given a user manual, a list of action names, variables, features, and a predefined \texttt{Appliance()} class in Python. Your task is to implement a \textbf{Simulator()} object as an instance of the predefined \texttt{Appliance()} object that models all action effects of the appliance.

\vspace{0.5em}
\textbf{Definition:} The \texttt{Simulator()} object inherits from \texttt{Appliance()} and implements three components:
\begin{enumerate}[label=(\arabic*)]
  \item \texttt{reset()} method that assigns: 
  \begin{itemize}
  \item \texttt{self.feature}, initialized as \texttt{simulator\_feature}.
  \item \texttt{self.variable\_x}, initialized from predefined variables.
  \item \texttt{self.variable\_input\_string}, \texttt{self.meta\_actions\_dict}, etc., if appliance includes number pads.
\end{itemize}
\item Action functions that define effects on variables and features. Valid action effects include:
\begin{itemize}
  \item Advance the current feature step or switch features by calling \texttt{self.feature.update\_progress(action\_name)}. Current feature and step index can be accessed by \texttt{self.feature.current\_value}.
  \item Get active variable via \texttt{self.get\_current\_variable(action\_name)}.
  \item Conditionally update variable value ranges or step size.
  \item Update variable value with \texttt{variable\_x.set\_current\_value()}, \texttt{self.assign\_variable\_to\_\_next(variable\_x)}, or \texttt{self.assign\_variable\_to\_\_prev(variable\_x)}.
\end{itemize}
\begin{tcolorbox}[colback=gray!5]
\begin{minted}[breaklines, breakanywhere]{python}
def press_a_button(self):
    self.feature.update_progress("press_a_button")
    current_feature = self.feature.current_value[0]
    variable = self.get_current_variable(action_name)
    if current_feature == "feature_a":
        variable.set_current_value("on")
    elif current_feature in ["feature_b", "feature_c"]:
        self.assign_variable_to_next(variable)
\end{minted}
\end{tcolorbox}
 \item \texttt{run\_action(action\_name, \dots)} is a wrapper that enforces global execution conditions before running an action. Specifically:
\begin{itemize}
  \item Prevents action execution when the appliance is locked or powered off, unless the action is to unlock or power on.
  \item Clears the input buffer (i.e., \texttt{self.variable\_input\_string}) if the action is unrelated to input via number pads.
  \item After passing precondition checks, invokes the corresponding action method to perform its effect.
\end{itemize}
\begin{tcolorbox}[colback=gray!5]
\begin{minted}[breaklines, breakanywhere]{python}
def run_action(self, action_name, execution_times=1, **kwargs):
    if action_name not in self.meta_actions_dict.values():
        self.variable_input_string.input_string = ""
    if self.variable_lock.get_current_value() == "locked" and "unlock" not in action_name:
        self.display = "child lock: locked"
        return self.display
    return super().run_action(action_name, execution_times, **kwargs)
\end{minted}
\end{tcolorbox}
\end{enumerate}
\vspace{0.5em}
\textbf{Example Output:}
\begin{tcolorbox}[colback=gray!5]
\begin{minted}[breaklines, breakanywhere]{python}
class Simulator(Appliance):

    def reset(self):
        self.feature = simulator_feature
        self.variable_clock_setting_hour = variable_clock_setting_hour
        self.variable_clock_setting_minute = variable_clock_setting_minute

    def press_clock_button(self): 
        ...

    def press_up_arrow_button(self):
        ...

    def press_down_arrow_button(self):
        ...

    def run_action(self, action_name, execution_times=1, **kwargs):
        ...
\end{minted}
\end{tcolorbox}
\vspace{0.5em}
\textbf{Other Valid Action Function Formats:}
\begin{enumerate}[label=(\arabic*)]
\item For \texttt{hold\_<element\_name>} actions, the duration needs to be included. 
\begin{tcolorbox}[colback=gray!5]
\begin{minted}[breaklines, breakanywhere]{python}
def press_and_hold_lock_button(self, duration=3):
    if duration >= 3:
        self.feature.update_progress("press_and_hold_lock_button")
        ...
\end{minted}
\end{tcolorbox}
\item If the action changes a program choice (e.g. \texttt{microwave menu}, \texttt{washing machine program}), sometimes the available program settings will change (e.g. \texttt{pizza} program requires setting \texttt{cooking\_weight}, but \texttt{soup} program requires setting \texttt{soup\_volume} (explained in \textit{extract variable}). Update the \texttt{variable\_program\_setting} accordingly.
\begin{tcolorbox}[colback=gray!5]
\begin{minted}[breaklines, breakanywhere]{python}
def press_menu_button(self):
    ...
    self.variable_program_setting = self.program_setting_dict[self.variable_program_index.get_current_value()]
\end{minted}
\end{tcolorbox}
\item If an action involves pressing number pads, follow this structure. 
\begin{itemize}
  \item Define a \texttt{press\_number\_button} method to model number pad action effects. Use this method to instantiate specific number pad actions.
  \begin{tcolorbox}[colback=gray!5]
\begin{minted}[breaklines, breakanywhere]{python}
# number pad action effects.
def press_number_button(self, action_name, digit):
    self.feature.update_progress(action_name)
    self.variable_input_string.add_digit(digit)
    variable = self.get_current_variable(action_name)
    value = self.process_input_string(current_feature, variable_name)
    variable.set_current_value(value)

# instantiate specific number pad actions.
def press_number_2_button(self): 
    self.press_number_button("press_number_2_button", "2")
\end{minted}
\end{tcolorbox}

  \item Define a \texttt{process\_input\_string} to convert inputs via number pads (e.g. \texttt{"1"}, \texttt{"6"}, \texttt{"0"}, \texttt{"0"}) to valid variable values (e.g. clock time of \texttt{"16:00"}).
    \begin{tcolorbox}[colback=gray!5]
  \begin{minted}[breaklines, breakanywhere]{python}
# converts time inputs of minute:second format to hour:minute:second format
def process_input_string(self, feature, variable_name):
    raw_input = self.variable_input_string.input_string
    if feature == "clock_setting" and variable_name == "variable_clock_time":
        time_string = "00" + str(raw_input).zfill(4)
        return f"{time_string[:2]}:{time_string[2:4]}:{time_string[4:]}"
\end{minted}
\end{tcolorbox}
  
  \item Define a \texttt{get\_original\_input} to convert target variable values (e.g. clock time of \texttt{"16:00"}) to required inputs via number pads (e.g. \texttt{"1"}, \texttt{"6"}, \texttt{"0"}, \texttt{"0"}). 
  \begin{tcolorbox}[colback=gray!5]
  \begin{minted}[breaklines, breakanywhere]{python}
# converts target time value of hour:minute:second format to required inputs of minute:second format
def get_original_input(self, goal, feature, variable_name):
    digits_only = ''.join(char for char in str(goal) if char.isdigit())
    if feature == "clock_setting" and variable_name == "variable_clock_time":
        return digits_only[2:].lstrip("0") or "0"
\end{minted}
\end{tcolorbox}

    \item In \texttt{reset()} method, add the following content.
      \begin{tcolorbox}[colback=gray!5]
  \begin{minted}[breaklines, breakanywhere]{python}
def reset(self):
    ... (the aforementioned variable assignments)
    self.variable_input_string = VariableInputString()
    self.meta_actions_dict = meta_actions_dict
    self.meta_actions_on_number = self.meta_actions_on_number
\end{minted}
\end{tcolorbox}
    
\end{itemize}
\end{enumerate}

\end{promptbox}

\newpage
\begin{promptbox}[label=prompt:generate_task_policy_and_goal_state]{Generate Task Policy and Goal State}
\small

You are given a user manual, a list of features, a list of variables, and a user instruction. Your task is to determine which features need to be executed and how variables should be set to fulfill the instruction.

\vspace{0.5em}
\textbf{Output Formats}
\begin{enumerate}[label=(\arabic*)]
\item a Python list \texttt{task\_policy} which defines the minimal ordered list of features needed to fulfill the user instruction. Use the following rules:
\begin{itemize}
  \item Every selected feature must set at least one variable required in the user instruction. 
  \item Exclude features whose variables are all covered by previous features.
  \item Include the feature to turn on the device and let it start running. 
\end{itemize}
\item a string \texttt{policy\_choice\_reason} that explains why each feature was selected. If multiple features are needed, explain what each contributes.
\item a \texttt{changing\_variables} list that includes all variables in the feature sequence, in order of appearance. Only include listed variables.
\item a \texttt{goal\_state = Simulator()} object. For each variable in \texttt{changing\_variables}, assign its target value following this structure:
\begin{itemize}
  \item Use \texttt{set\_current\_value()} for direct assignment.
  \item Use \texttt{set\_value\_range()} or \texttt{set\_step\_value()} if the variable's default configuration changes.
  \item Do not modify variable names. Use the exact names from \texttt{changing\_variables}.
  \item For \texttt{ContinuousVariable} and \texttt{TimeVariable}, add a Python comment indicating unit (e.g., seconds, minutes, hours). \vspace{0.5em}
\end{itemize}
\end{enumerate}
\vspace{0.5em}
\textbf{Example Output:}
\begin{tcolorbox}[colback=gray!5]
\begin{minted}[breaklines, breakanywhere]{python}
# User Instruction: Defrost chicken meat for 5 minutes at 50% power in 3 hours time. 
task_policy = ["cook", "preset", "start"]
policy_choice_reason = "Firstly adjust cook settings then set preset hours."
changing_variables = ["variable_microwave_cooking_power", "variable_microwave_cooking_time", "variable_preset_time", "variable_start"]
goal_state = Simulator()
goal_state.variable_microwave_cooking_power.set_current_value("P50")
goal_state.variable_microwave_cooking_time.set_current_value("00:05:00") # 5 minutes
goal_state.variable_preset_time.set_current_value(3)  # hour
goal_state.variable_start.set_current_value("on")
\end{minted}
\end{tcolorbox}

\vspace{0.5em}
\textbf{Handle Program Choices:}
An appliance may allow choosing different programs (e.g. microwave menu, washing machine program), and each program has different settings (e.g. \texttt{pizza} program requires setting \texttt{cooking\_weight}, but \texttt{soup} program requires setting \texttt{soup\_volume} (explained in \textit{extract variable}). In this case, \texttt{variable\_program\_setting} will be initialized with \texttt{None} in \texttt{reset()}. Therefore in \texttt{goal\_state}, firstly assign it to an existing defined variable (e.g., from a mapping dictionary), and set its value accordingly. 
\begin{tcolorbox}[colback=gray!5]
\begin{minted}[breaklines, breakanywhere]{python}
# Given variables
variable_program_index = DiscreteVariable(["popcorn", "pizza",
"soup"], "popcorn"),
variable_program_setting = None

variable_program_setting_popcorn = DiscreteVariable(["1 cup",
"2 cup"], "1 cup"),
variable_program_setting_pizza = DiscreteVariable(["250g",
"350g", "450g"], "250g"),
variable_program_setting_soup = DiscreteVariable(["200ml",
"300ml", "400ml"], "200ml"),

program_setting_dict = {
"popcorn": variable_program_setting_popcorn,
"pizza": variable_program_setting_pizza,
"soup": variable_program_setting_soup
}

# Given feature
feature_list["set_program"] = [
    {"step": 1, "actions": ["press_program_button"], "variable": "variable_program_index"},
    {"step": 2, "actions": ["press_up_arrow_button", "press_down_arrow_button"], "variable": "variable_program_setting"}
]

# User Instruction: Set the microwave to cook 1 cup of popcorn...
task_policy = ["set_program"]
policy_choice_reason = "This feature contains variable_program_index and variable_program_setting". 
changing_variables = ["variable_program_index", "variable_program_setting"]
goal_state = Simulator()
goal_state.variable_program_index.set_current_value("popcorn")
goal_state.variable_program_setting = variable_program_setting_popcorn
goal_state.variable_program_setting.set_current_value("1 cup")
\end{minted}
\end{tcolorbox}

\end{promptbox}

\newpage
\begin{promptbox}[label=prompt:compare_goal_state_with_feedback]{Compare Goal State with Feedback}
\small

You are given the appliance model, together with two strings in the format \texttt{variable\_name: variable\_value}, representing the goal state and the real-world feedback, respectively. Your task is to determine whether the feedback indicates the goal is reached.

\vspace{0.5em}
\textbf{Comparison Rules:}
\begin{enumerate}[label=(\arabic*)]
    \item Allow equivalent variable-value meaning.
    \textit{E.g.} \texttt{variable\_menu = "Popcorn"} vs. \texttt{mode\_popcorn = "on"} $\Rightarrow$ \texttt{True}; \texttt{variable\_power = "On"} vs. \texttt{variable\_on\_off = "On"} $\Rightarrow$ \texttt{True}

    \item If values contain both numbers and text, remove text and compare numbers. Ignore casing or formatting if numerically identical.
    \textit{E.g.} \texttt{"0g"} vs. \texttt{"0"} $\Rightarrow$ \texttt{True}; \texttt{"100cm"} vs. \texttt{"100"} $\Rightarrow$ \texttt{True}; \texttt{"1 cup"} vs. \texttt{"1 serving"} $\Rightarrow$ \texttt{True}

    \item Ensure the match is the closest in the value range.
    \textit{E.g.} \texttt{program="wash"} v.s \texttt{program="wash, dry"}, both values exist in value range $\Rightarrow$ \texttt{False}

\end{enumerate}

\vspace{0.5em}
\textbf{Output Format:}
\begin{itemize}
    \item \texttt{reason}: a string explaining your judgment.
    \item \texttt{goal\_reached}: either \texttt{True} or \texttt{False}.
\end{itemize}

\vspace{0.5em}
\textbf{Example Output:}
\begin{tcolorbox}[colback=gray!5]
\begin{minted}[breaklines, breakanywhere]{python}
# goal: popcorn setting = 100g;
# feedback: popcorn: 100
reason = "Both values represent 100g, ignoring unit suffix."
goal_reached = True
\end{minted}
\end{tcolorbox}

\end{promptbox}

\begin{promptbox}[label=prompt:diagnose_incorrect_variable_definition]{Diagnose Incorrect Variable Definition}
\small

You are given:

\begin{itemize}
    \item A list of defined variable names in the appliance model.
    \item A variable name \texttt{variable\_x} suspected to be incorrectly defined.
    \item A full step-by-step execution record starting from the first observed change in that variable's value. Each record includes the action taken and the observed result in the format: \texttt{variable\_name = variable\_value}.
\end{itemize}

\vspace{0.5em}
\textbf{Your Tasks:}
\begin{enumerate}[label=(\arabic*)]
    \item \textbf{Identify the root variable:}
    \begin{itemize}
        \item Match the observed variable name to the closest name in the given variable list. If the mismatch is caused by this variable itself, return that name as \texttt{variable\_name}.
        \item If the variable is conditioned on a program choice (e.g., \texttt{variable\_program\_setting}), and the mismatch is due to a sub-variable (e.g., \texttt{variable\_program\_setting\_popcorn}), return the name of the sub-variable.
    \end{itemize}

    \item \textbf{Determine if the variable is continuous:}  
    \begin{itemize}
        \item Return \texttt{variable\_is\_continuous = True} if the values are numeric and increase/decrease regularly.
        \item Else return \texttt{variable\_is\_continuous = False}.
    \end{itemize}

    \item \textbf{Extract the variable values as a list:}  
    \begin{itemize}
        \item Extract all values of the observed variable in order from the record.
        \item Store them in \texttt{record\_sequence}.
        \item Use \texttt{int}/\texttt{float} for continuous variables and \texttt{str} for discrete ones.
    \end{itemize}
\end{enumerate}

\vspace{0.5em}
\textbf{Output Format:}
Return the following Python variables:
\begin{itemize}
    \item \texttt{variable\_name}
    \item \texttt{variable\_is\_continuous}
    \item \texttt{record\_sequence}
\end{itemize}

\vspace{0.5em}
\textbf{Example:}
\begin{tcolorbox}[breakable, colback=gray!5]
\begin{minted}[breaklines, breakanywhere]{python}
# inputs given
defined_variables = [
    "variable_wash_time",
    "variable_spin_speed",
    "variable_temperature"
]

execution_record = [
  {step_index: 1, action: ("turn_dial", 1), observation: wash_time = 6},
  {step_index: 2, action: ("turn_dial", 1), observation: wash_time = 9},
  {step_index: 3, action: ("turn_dial", 1), observation: wash_time = 12},
  ...
]

# Expected Output
variable_name = "variable_wash_time"
variable_is_continuous = True
record_sequence = [6, 9, 12, ...]
\end{minted}
\end{tcolorbox}

\end{promptbox}

\begin{promptbox}[label=prompt:update_variable_definition]{Update Variable Definition from Observed Values}
\small

You are given the following inputs:
\begin{itemize}
    \item \texttt{variable\_name}: the variable that has been confirmed to be incorrectly defined.
    \item \texttt{variable\_is\_continuous}: whether the variable is continuous or discrete.
    \item \texttt{record\_sequence}: the list of observed values of the variable over time.
    \item The current implementation of the variable.
    \item The user manual and a guide for valid variable definitions.
\end{itemize}

\vspace{0.5em}
\textbf{Your Task:}
Update the variable definition by modifying its current value, value range, step size, or value order to match all values in \texttt{record\_sequence}.

\vspace{0.5em}
\textbf{Instructions:}
\begin{enumerate}[label=(\arabic*)]
    \item \textbf{Paste the reasoning trace:} Insert the provided \texttt{record\_sequence} as Python comments to justify your updates.

    \item \textbf{Update the variable:} Modify the definition of the chosen variable to match observed behavior. Keep the same name. Valid modifications include:
    \begin{enumerate}[label=(\alph*)]
        \item \textbf{Change variable type} according to observation.  
        \item \textbf{Change current value} to match with the last observed value.
        \item \textbf{Adjust value range or step size} if the record shows regular repetition. Use piecewise ranges if steps skip sections.
        \item \textbf{Change value order} for discrete variables if observed cycling order differs.
    \end{enumerate}
    
    \item \textbf{Copy related data structures:} If the variable is part of a program-conditioned setting (e.g., \texttt{variable\_program\_setting}, explained in \textit{extract variables}), also update the program dictionary:
    \begin{tcolorbox}[breakable, colback=gray!5]
    \begin{minted}{python}
program_setting_dict["menu_x"] = variable_x
    \end{minted}
    \end{tcolorbox}

    \item \textbf{Align with real-world units.} For example, if feedback is in \texttt{cm}, don’t define value ranges in \texttt{m}. For continuous variables representing time or weight, indicate the unit in a Python comment (e.g., seconds, minutes, grams).
\end{enumerate}

\vspace{0.5em}
\textbf{Example Output:}
\begin{tcolorbox}[breakable, colback=gray!5]
\begin{minted}[breaklines, breakanywhere]{python}
# given inputs
variable_name = "variable_program_setting_popcorn"
variable_is_continuous = True
record_sequence = [0, 100, 200, 300, 400, 0]

# record_sequence = [0, 100, 200, 300, 400, 0]
# Step size = 100; values loop back to 0
# Range spans 0 to 400 with step 100
variable_program_setting_popcorn = ContinuousVariable(
    value_ranges_steps=[(0, 400, 100)],
    current_value=0
) # in grams
program_setting_dict["popcorn"] = variable_program_setting_popcorn
\end{minted}
\end{tcolorbox}

\end{promptbox}

\begin{promptbox}[label=prompt:update_appliance_model]{Update Appliance Model After Updating Variable}
\small

You are given:
\begin{itemize}
    \item The original simulator implementation.
    \item The incorrect variable name, \texttt{variable\_x}.
    \item The corrected variable definition.
\end{itemize}

\vspace{0.5em}
\textbf{Your Task:}  
Update the \texttt{Simulator()} class so that all references to \texttt{variable\_x} reflect its corrected definition.

\vspace{0.5em}
\textbf{Instructions:}
\begin{enumerate}[label=(\arabic*)]
    \item For \texttt{Simulator()}, edit only affected action methods. Keep unrelated parts of the simulator unchanged. Do not modify or omit the \texttt{reset()} method. 

    \item Exclude code outside \texttt{Simulator()}, such as class definitions (\texttt{Appliance()}, \texttt{Variable()}), variables and \texttt{simulator\_feature}. 

\end{enumerate}

\vspace{0.5em}
\textbf{Example Output:}
\begin{tcolorbox}[breakable, colback=gray!5]
\begin{minted}[breaklines, breakanywhere]{python}
# variable_power was changed from ContinuousVariable to DiscreteVariable. The valid value ranges change from float (e.g. 100) to string (e.g. "100"). 
class Simulator(Appliance):
    def reset(self):
        ...

    def press_start_button(self):
        self.feature.update_progress("press_start_button")
        current_feature = self.feature.current_value[0]
        if current_feature == "speed_cook":
            self.assign_variable_to_next(self.variable_cooking_time)
            # updated line
            self.variable_power.set_current_value("100") 
\end{minted}
\end{tcolorbox}

\end{promptbox}

\begin{promptbox}[label=prompt:update_goal]{Update Goal Value After Variable Definition Change}
\small

You are given a user instruction, an appliance model, a goal state object 
\begin{itemize}
    \item a user instruction.
    \item the implemented appliance model, i.e., a \texttt{Simulator()} object.
    \item A \texttt{goal\_state = Simulator()} object specifying target variable values that achieves the instruction. 
    \item The updated variable name, \texttt{variable\_x}.
    \item A goal-setting guide for reference.
\end{itemize}

\vspace{0.5em}
\textbf{Your Task:}
Update the goal value of \texttt{variable\_x} in the goal state to match the new definition.

\vspace{0.5em}
\textbf{Instructions:}
\begin{enumerate}[label=(\arabic*)]
    \item Ensure the new value assignment aligns with both the the updated definition \texttt{variable\_x} and the user instruction. 
    \item Do not rename \texttt{variable\_x}. Do not modify any other variables in the goal state.
    \item Do not return any other content (e.g., comments, reasoning, variable definitions, or unrelated goal assignments).
\end{enumerate}

\vspace{0.5em}
\textbf{Output Format:}
A single line of valid Python code that updates \texttt{goal\_state.variable\_x} to the correct value.

\vspace{0.5em}
\textbf{Example Output:}
\begin{tcolorbox}[colback=gray!5]
\begin{minted}{python}
# updated timer to ContinuousVariable, previously was DiscreteVariable
goal_state.variable_microwave_timer.set_current_value(3) # minutes
\end{minted}
\end{tcolorbox}

\end{promptbox}

\begin{promptbox}[label=prompt:check_bbox_contain_control_panel_elements]{Check if Bounding Box Contains Control Panel Element}
\small

\textbf{Task:}  
Given an image labeled with a bounding box, determine whether the bounding box contains a control panel element.

\vspace{0.5em}
\textbf{Definition of Control Panel Element:}
Control panel elements include:
\begin{itemize}
    \item Physical components: buttons, dials.
    \item Soft pads: labels printed directly on the control surface that respond to touch input. These labels might include printed symbols and icons, such as: "+", "-", "start", "on/off", and numeric digits.
\end{itemize}

\vspace{0.5em}
\textbf{Instructions:}
\begin{enumerate}[label=(\arabic*)]
    \item Review the region circled by the bounding box.
    \item If the bounding box contains any of the valid elements listed above, reply with \texttt{"Yes"}. Otherwise reply with \texttt{"No"}.
    \item In both cases, provide a reason by naming the object being circled by the red bounding box.
\end{enumerate}

\vspace{0.5em}
\textbf{Output Format:}
\begin{tcolorbox}[colback=gray!5]
\begin{minted}{text}
Yes
Reason: The red box surrounds the "+" symbol on the soft pad region.
\end{minted}
\end{tcolorbox}

\end{promptbox}

\begin{promptbox}[label=prompt:map_bbox_to_names]{Map Bounding boxes to Control Panel Element Names}
\small

You are given:
\begin{itemize}
    \item A list of control panel element names including buttons, dials, and soft-labeled pads.
    \item Three images:
    \begin{enumerate}[label=(\arabic*)]
        \item Full view of the control panel.
        \item Zoomed-in region with a red bounding box and several green bounding boxes.
        \item Same zoomed-in region without bounding boxes.
    \end{enumerate}
    \item A \texttt{bounding\_box\_index} referring to the red box.
\end{itemize}

\vspace{0.5em}
\textbf{Your Task:}
\begin{enumerate}[label=(\arabic*)]
    \item Determine whether the red bounding box encloses a listed control element. Be lenient: if the red box contains any label, symbol, or visible control region, attempt to match. If multiple names match the red box, include them all.
    \begin{itemize}
        \item For \textbf{dials}: Only bounding boxes covering the knob are valid. Ignore labels around the dial.
        \item For \textbf{buttons}: Only bounding boxes that cover the physical, pressable area are valid. Boxes that only enclose external labels are invalid.
        \item For \textbf{soft-labeled pads}: If the label itself is the interactive surface (i.e., no visible border or physical button), bounding boxes over the label region are valid.
    \end{itemize}

    \item If (1) is true, check if the red box is a better match than any green box for the same element. 
    \begin{itemize}
        \item It is okay for red box to partially enclose the object.
        \item If red box is clearer or more precise than all green boxes, accept it as the match. 
    \end{itemize}
\end{enumerate}

\vspace{0.5em}
\textbf{Output Format:}
\begin{itemize}
    \item If both conditions are met, output the matched control element(s) in format below. Use exact names from the provided list. 
    \begin{tcolorbox}[breakable, colback=gray!5]
\begin{minted}{text}
<control_element_name> : <index>
<control_element_name> : <index>
...
\end{minted}
\end{tcolorbox}

    \item If no valid match is found, output \texttt{None}.
\end{itemize}

\vspace{0.5em}
\textbf{Example Output:}
\begin{tcolorbox}[breakable, colback=gray!5]
\begin{minted}[breaklines, breakanywhere]{python}
temperature dial : 1
power dial: 2
None
temperature dial: 3
power dial: 3
\end{minted}
\end{tcolorbox}

\end{promptbox}

\begin{promptbox}[label=prompt:remove_duplicate_bbox]{Remove Duplicate Bounding Boxes for Control Panel Elements}
\small

You are given an \texttt{appliance\_type}, which contains a \texttt{control\_panel\_element\_name}. Control panel elements are components responsible for operating the appliance, such as buttons, dials and soft touch pads. 
\vspace{0.5em}
You are given:
\begin{itemize}
    \item A photo of the appliance to identify \texttt{control\_panel\_element\_name}.
    \item A sequence of images showing bounding box options around potential regions for \texttt{control\_panel\_element\_name}. Each box has a visible \texttt{index} at its bottom-right corner.
\end{itemize}

\vspace{0.5em}
\textbf{Your Task:}  
Select \textbf{one} bounding box index that best matches the \texttt{control\_panel\_element\_name}. If none of the bounding boxes is valid, return \texttt{response\_index = -1}.

\vspace{0.5em}
\textbf{Selection Criteria:}
\begin{itemize}
    \item \textbf{Dial:} Choose the bounding box that covers the \textit{knob}. Ignore boxes that only include labeling or surrounding text.
    
    \item \textbf{Button:}
    \begin{itemize}
        \item If the label is printed directly on the button, a box selecting either the full button or label area is valid, even if the coverage is partial.
        \item If the label is outside a physical button, select the bounding box around the physical (extruded) button, not just the label.
    \end{itemize}
    
    \item \textbf{Soft Pad:} When the label text or icon \textit{is} the button (i.e., not physically extruded), select the box that covers any part of that label or symbol.
\end{itemize}

\vspace{0.5em}
\textbf{Output Format:}
Return two variables in Python format:

\begin{tcolorbox}[breakable, colback=gray!5]
\begin{minted}[breaklines, breakanywhere]{python}
response_index = 3
response_reason = "The bounding box covers the soft pad label text of the button."
\end{minted}
\end{tcolorbox}

If no bounding box fits the criteria:
\begin{tcolorbox}[breakable, colback=gray!5]
\begin{minted}[breaklines, breakanywhere]{python}
response_index = -1
response_reason = "None of the boxes select the physical button or label. The target is a circular dial knob near the bottom left corner."
\end{minted}
\end{tcolorbox}

\end{promptbox}

\begin{promptbox}[label=prompt:ground_actions]{Ground Actions}
\small

You are given a list of action names and a list of control panel element names. Your task is to ground each action to a control panel element name and a valid action type. Valid action types include \texttt{press}, \texttt{hold}, \texttt{turn\_dial\_clockwise}, \texttt{turn\_dial\_anti\_clockwise}.

\vspace{0.5em}
\textbf{Output Format:}  
Return a Python list of dictionaries. Each dictionary contains a grounded action, with the following keys:
\begin{enumerate}[label=(\arabic*)]
    \item \texttt{"action"}: a string from the given action list (e.g., \texttt{"press\_max\_crisp\_button"}).
    \item \texttt{"bbox\_label"}: a list of strings from the given control element names. 
    \begin{itemize}
        \item For standard actions, this is a single-element list (e.g., \texttt{["max\_crisp\_button"]}).
        \item For simultaneous actions (e.g., \texttt{hold\_wash\_button\_and\_rinse\_button}), include both elements (e.g., \texttt{["wash\_button", "rinse\_button"]}).
    \end{itemize}
    \item \texttt{"action\_type"}: inferred from the action name string using the following rules:
    \begin{itemize}
        \item Contains \texttt{"hold"} $\Rightarrow$ \texttt{"hold\_button"}
        \item Contains \texttt{"press"} $\Rightarrow$ \texttt{"press"}
        \item Contains \texttt{"turn\_dial\_clockwise"} $\Rightarrow$ \texttt{"turn\_dial\_clockwise"}
        \item Contains \texttt{"turn\_dial\_anti\_clockwise"} $\Rightarrow$ \texttt{"turn\_dial\_anti\_clockwise"}
    \end{itemize}
\end{enumerate}

\vspace{0.5em}
\textbf{Example Output:}
\begin{tcolorbox}[breakable, colback=gray!5]
\begin{minted}[breaklines, breakanywhere]{python}
[
    {
        "action": "press_max_crisp_button",
        "bbox_label": ["max_crisp_button"],
        "action_type": "press_button"
    },
    {
        "action": "press_and_hold_cancel_button_and_stop_button",
        "bbox_label": ["cancel_button", "stop_button"],
        "action_type": "press_and_hold_button"
    }
]
\end{minted}
\end{tcolorbox}

\end{promptbox}

\begin{promptbox}[label=prompt:visual_feedback_parsing]{Visual Feedback Parsing}
\small

You are given:
\begin{itemize}
    \item A user command describing the task.
    \item The most recent action applied and the target variable being adjusted.
    \item The valid value range of the target variable.
    \item An image of the appliance control panel after the action.
    \item Relevant user manual text describing the display panel.
\end{itemize}

\vspace{0.5em}
\textbf{Your Task:}
\begin{itemize}
    \item Interpret the display image to infer the current appliance state, especially the value of the target variable.
    \item Use the user manual to explain display symbols if needed.
\end{itemize}

\vspace{0.5em}
\textbf{Output Format:} 
\begin{tcolorbox}[colback=gray!5]
\begin{minted}[breaklines]{python}
variable_description = "<Concise interpretation of the current state, focusing on the target variable.>"
\end{minted}
\end{tcolorbox}

\vspace{0.5em}
\textbf{Example:}
\begin{tcolorbox}[colback=gray!5]
\begin{minted}[breaklines]{python}
# Task: Set temperature to 98°C. Action: 'press_temp_clean_button'.
# Display shows a triangle under 85°C.

variable_description = "The triangle under '85°C' indicates the current selection. variable_temperature = 85."
\end{minted}
\end{tcolorbox}
\end{promptbox}

\end{document}